\pdfoutput=1

\documentclass[11pt]{article}

\usepackage[preprint]{acl}
\usepackage{booktabs}
\usepackage{multirow}
\usepackage{times}
\usepackage{latexsym}
\usepackage{booktabs}
\usepackage{tabularx}
\usepackage{makecell}
\usepackage{framed}

\usepackage{mdframed}
\usepackage{todonotes}
\usepackage{hyperref}

\newcommand{\sectioncolor}{violet}

\usepackage[T1]{fontenc}

\usepackage[utf8]{inputenc}

\usepackage{microtype}
\usepackage[most]{tcolorbox}

\usepackage{inconsolata}
\usepackage{xcolor}

\usepackage{multirow}

\usepackage{graphicx}
\usepackage{float}
\usepackage{amsmath, amssymb}

\newcommand\jee{mmJEE-Eval}

\definecolor{lightred}{RGB}{255,200,200}
\definecolor{lightgreen}{RGB}{200,255,200}
\definecolor{lightyellow}{RGB}{255,255,200}
\definecolor{lightblue}{RGB}{200,230,255}
\definecolor{darkred}{RGB}{180,0,0}

\newcommand{\wrongtag}[1]{\textbf{\textcolor{darkred}{#1}}}

\newtcolorbox{promptbox}[1][]{
    colback=blue!5!white,
    colframe=blue!75!black,
    fonttitle=\bfseries,
    title=#1,
    arc=3pt,
    outer arc=3pt,
    boxrule=0.5pt,
    left=6pt,
    right=6pt,
    top=6pt,
    bottom=6pt
}

\title{mmJEE-Eval: A Bilingual Multimodal Benchmark for Evaluating Scientific Reasoning in Vision-Language Models}

\author{Arka Mukherjee\thanks{Work done while at IIT Bhubaneswar} \\ Kalinga Institute of  \\ Industrial Technology (KIIT) \\ \texttt{arka.mukherjee078@gmail.com} 
        \And  
         Shreya Ghosh \\ Indian Institute of Technology \\ (IIT), Bhubaneswar \\ \texttt{shreya@iitbbs.ac.in} 
}

\begin{document}
\maketitle

\begin{abstract}

Contemporary vision-language models (VLMs) perform well on existing multimodal reasoning benchmarks (78-85\% accuracy on MMMU, MathVista). Yet, these results fail to sufficiently distinguish true scientific reasoning articulation capabilities from pattern-matching. To address this gap, we introduce \textbf{mmJEE-Eval}, a multimodal bilingual (English and Hindi) benchmark comprising 1,460 questions from India's JEE Advanced examination (2019-2025) spanning pre-college Physics, Chemistry, and Mathematics domains. Our evaluation of 17 state-of-the-art models reveals that while frontier VLMs (GPT-5, Gemini 2.5 Pro/Flash) achieve 77-84\% accuracy on held-out 2025 questions, open-source models plateau at 37-45\% despite scaling to 400B parameters, a significant difference not observed on existing benchmarks. While closed frontiers from Google and OpenAI show high problem-solving accuracies (up to 100\% pass@3 scores), they fully collapse when the reasoning load is increased meta-cognitively (GPT-5 fixes just 5.2\% errors). Systematic ablations show mmJEE-Eval's difficulty stems from complexity and reasoning depth rather than memorization. Effectively, our benchmark segregates superior training and reasoning methodologies where alternatives fail. We publicly release our code and data: \url{https://mmjee-eval.github.io}
\end{abstract}

\section{Introduction}

Recent advances in vision–language models (VLMs) have produced systems that perform impressively on broad multimodal leaderboards. To evaluate these systems, numerous benchmarks exist: MMMU~\cite{yue2024mmmumassivemultidisciplinemultimodal}, MathVista~\cite{lu2024mathvista}, MMMU-Pro~\cite{yue-etal-2025-mmmu}, and CharXiv~\cite{NEURIPS2024_cdf6f8e9} test multimodal reasoning across diverse domains.

\begin{figure}[H]
   \centering
   \includegraphics[width=\columnwidth]{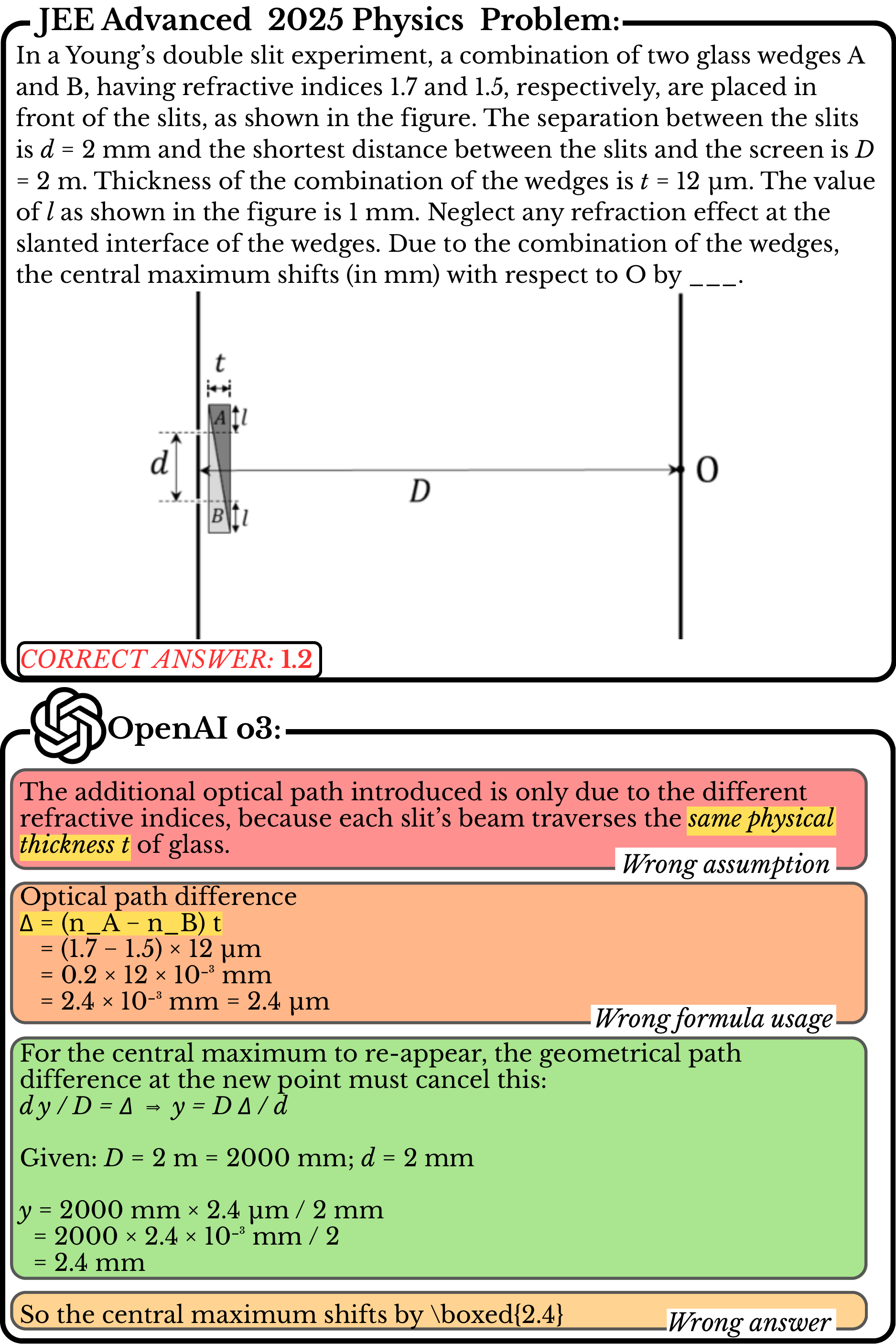}
   \caption{Example problem and response from \jee. Despite mathematical correctness, the model incorrectly assumes uniform thickness ("same physical thickness $t$ of glass") when the figure clearly shows wedge-shaped glass pieces with varying thickness. This multimodal reasoning failure demonstrates the multiple dimensions our proposed benchmark tests.}
   \label{fig:model_response_example}
\end{figure}

These benchmarks show that recent open-source VLMs are generally within single-digit to low-double-digit points of leading closed systems, and occasionally surpass them. On MMMU, Qwen3 VL 235B (78.7\%) and Llama 4 Maverick (73.4\%) trail Gemini 2.5 Pro (84.0\%) and GPT-5 (84.2\%) by modest margins. MMMU-Pro shows a slightly wider trend (68.1\% vs 78.4\%). However, on MathVista, Qwen3 VL 235B achieves 84.9\%, edging out Gemini 2.5 Pro's 84.6\% and GPT-5's 82.7\%. Taken together, these results suggest that while divergence slightly widens on multimodal, multi-step compositional reasoning, the gap between open and closed models has narrowed.

Recent efforts like Humanity's Last Exam (HLE)~\cite{phan2025humanitysexam} show that pure difficulty widens the difference. While GPT-5 (25.3\%), Gemini 2.5 Pro (21.6\%), and o3 (20.3\%) cluster below 26\%, open models score 5-12\%. Yet such uniformly low scores provide limited insight into \textit{why} models struggle or which capabilities remain deficient.

To address this, we introduce mmJEE-Eval, a multimodal bilingual benchmark built from seven years (2019-2025) of India's JEE Advanced examination, a rigorous high school-level entrance test for India's premier technical institutions, the IITs. JEE Advanced provides several properties valuable for evaluation: exam-style constraints including negative marking for incorrect responses; native bilingual presentation with questions appearing in both English and Hindi (as in the original exam booklets); integrated STEM reasoning across physics, chemistry, and mathematics; and contamination avoidance through annual new questions, provided with our 2025 held-out set~(Section~\ref{results:contamination}). While individual properties appear in prior work, their combination enables systematic diagnosis of failure modes. Figure~\ref{fig:model_response_example} illustrates such a multimodal reasoning failure from OpenAI o3, where the model is mathematically correct but misinterprets the visual setup.

To further understand such multimodal reasoning gaps, we answer the following research questions in this study:

\textbf{RQ1}: Does mmJEE-Eval’s exam-style multimodal evaluation reveal additional VLM capability gaps compared to standard benchmarks?

\textbf{RQ2}: Are VLMs resilient to semantically equivalent input language perturbations?

\textbf{RQ3}: Can contemporary VLMs detect and fix errors in their reasoning chains?

\textbf{RQ4}: What factors make mmJEE-Eval difficult?

Despite covering pre-college material, mmJEE-Eval reveals dramatic performance gaps absent in prior benchmarks. Through comprehensive evaluation of 17 VLMs (Section~\ref{results:acc}), we show Llama 4 Maverick scores 32.8\% while Gemini 2.5 Flash achieves 83.3\% on realistic exam marking (a 50.5-point gap), far larger than on MMMU, MMMU-Pro, and MathVista. This disparity suggests that \textit{problem complexity alone does not limit reasoning capabilities}. Through rigorous experiments and ablations, we attribute this gap to an absence of joint multilingual understanding, complex visual reasoning, and cross-domain concept integration mastery (Section~\ref{results:ablations}).

Beyond measuring problem-solving accuracy, we also probe whether models can detect and correct their own reasoning errors (Section~\ref{results:meta}). Here, we find universal metacognitive failure: both open and closed models detect errors in 21-73\% of cases but correct only 1.1-5.2\% of them, far below the 30-40\% typical pass@k gains. This suggests contemporary VLMs lack meta-cognitive capabilities for self-correction, even when they eventually get problems correct, given multiple attempts. 

\section{Related Work}
\label{sec:rel-work}

\subsection{General Reasoning Datasets}

Mathematical reasoning benchmarks have been the standard for evaluating LLM capabilities. MATH~\cite{hendrycks2021measuring} introduced 12,500 challenging competition mathematics problems from AMC, AIME, and Mathematical Olympiads. GSM8K~\cite{cobbe2021training} focused on 8,500 linguistically diverse grade school math word problems. These early datasets are English only, failing to capture the global scale of modern VLMs.

Recent efforts have scaled both in sophistication and LLM trends, such as instruction tuning. MathInstruct~\cite{yue2023mammoth} compiled 260,000 problems from 13 math datasets with intermediate rationales using hybrid Program-of-Thought and Chain-of-Thought approaches. MetaMathQA~\cite{yu2024metamath} bootstrapped 395,000 mathematical questions using GPT-3.5 Turbo-generated rephrasings to address the performance gap between open-source and closed-source models. However, these datasets primarily focus on computational mathematics rather than the advanced conceptual reasoning required in competitive examinations.

\subsection{Multilingual Reasoning Datasets}

Multilingual mathematical reasoning has emerged as a critical research direction, though most efforts rely on translation-based approaches. Notable datasets include X-CSQA and X-CODAH~\cite{lin-etal-2021-common}, which automatically translated English commonsense reasoning datasets to 16 languages using DeepL Pro. XCOPA~\cite{ponti-etal-2020-xcopa} manually translated commonsense reasoning tasks to 11 typologically diverse languages.

More sophisticated approaches include mCSQA~\cite{sakai2024mcsqa}, which proposed efficient methods to create multilingual NLU datasets using multilingual generative language models for cross-linguistic commonsense understanding analysis. Further, TyDiQA~\cite{10.1162/tacl_a_00317} collected question answering data directly in 11 typologically diverse languages without translation, setting the gold standard for multilingual evaluation.

\section{Data Collection}
\label{sec:methodology}

\jee~is collected and verified through a rigorous human annotation process to ensure high-quality data that is sufficiently challenging for state-of-the-art VLMs. Following \citet{arora-etal-2023-llms}, we collect JEE Advanced questions from the official archive\footnote{\url{https://jeeadv.ac.in/archive.html}}. After preprocessing, our multimodal dataset comprises 1,460 questions spanning seven editions of the exam (2019--2025).
 
\subsection{Dataset Collection and Annotation Process}

Given the inherent complexity of extracting structured information from multimodal PDF documents, our initial attempts using automated PDF parsing tools proved inadequate. We also evaluated automated bounding box detection techniques, including DART~\cite{Xin_2024}, which failed to handle challenging question types such as matching problems and exhibited poor generalization across years due to formatting variations in the question papers.


To ensure accurate collection of multimodal data from JEE Advanced PDFs, we developed custom manual annotation tools with multiple verification steps for both question and solution collection (detailed in Appendix~\ref{sec:manual-collection-software}).

Our annotation software automatically downloads all question papers and presents them to annotators in a page-by-page interface. The annotation task requires drawing precise bounding boxes around individual questions. We manually processed all 24 question papers, with particular attention to ensuring question images remain identifiable and complete. This process yielded an initial collection of 1,476 question images.

Collecting accurate ground-truth answers for all questions in our dataset is essential for proper evaluation. However, we discovered that not all JEE Advanced question papers are accompanied by official answer keys. Given that comprehensive solutions have been archived by third-party educational websites, we implemented a multi-source verification approach by collecting answer keys from five independent nation-wide popular coaching institutes: Aakash\footnote{\url{https://www.aakash.ac.in/jee-advanced-previous-year-question-papers}}, FIITJEE\footnote{\url{https://old.fiitjee.com/downloads/jee-advanced-(earlier-iit-jee)-solutions}}, Allen Kota\footnote{\url{https://allen.in/jee-advanced/previous-year-papers-with-solutions}}, Motion Education\footnote{\url{https://motion.ac.in/blog/jee-advanced-previous-year-question-papers}}, and Resonance\footnote{\url{https://www.resonance.ac.in/answer-key-solutions/JEE-Advanced.aspx}}. 

These institutes annually publish detailed solutions for all JEE Advanced papers to assist students in performance assessment, with all analyses carefully archived on their respective websites for current aspirants. We collect six independent answers per question and employ majority voting to determine the ground-truth solution. Solutions receiving greater than 60\% consensus are accepted into our dataset. Questions failing to achieve a majority consensus are flagged for manual verification. This discrepancy most commonly occurred in numerical-type questions, where official answer keys accept ranges of values while independent solutions provide singular values or narrower ranges. Only 30 such cases were recorded out of 1,476, for which we deferred to the official answer key when available. Table~\ref{tab:consensus-level} presents an analysis of confidence levels recorded in the answer collection process.

\begin{table}[h]
\centering
\begin{tabular}{c|c|c}
\hline
\textbf{Consensus Level} & \textbf{Questions} & \textbf{Percentage} \\
\hline
100\%        & 1326 & 90.8 \\
80\% -- 99.99\%  & 68 & 4.7 \\
60\% -- 79.99\% & 36 & 2.4 \\
50\% -- 59.99\%  & 26 & 1.8 \\
\textless 50\%     & 4    & 0.3  \\
\hline
\end{tabular}
\caption{Consensus level statistics for manually collected answers for \jee from six independent sources.}
\label{tab:consensus-level}
\end{table}

\subsection{Data Statistics}
\label{sec:data-stat}

Finally, \jee~comprises 1,460 questions spanning seven years (2019-2025). As detailed in Table~\ref{tab:yearwise-subject-questiontype}, the dataset maintains a balanced distribution across all three core subjects: Chemistry (492 questions, 33.7\%), Mathematics (492 questions, 33.7\%), and Physics (476 questions, 32.6\%).  

\begin{table}[h!]

\centering
\resizebox{\columnwidth}{!}{%
\setlength{\tabcolsep}{3pt}
\renewcommand{\arraystretch}{1.1}
\begin{tabular}{c|ccc|cccc|c}
\hline
\multirow{2}{*}{\textbf{Year}} & \multicolumn{3}{c|}{\textbf{Subjects}} & \multicolumn{4}{c|}{\textbf{Question Types}} & \multirow{2}{*}{\textbf{Total}} \\
\cline{2-4} \cline{5-8}
 & \textbf{Chem} & \textbf{Math} & \textbf{Phys} & \textbf{Multi} & \textbf{Single} & \textbf{Num.} & \textbf{Match.} & \\
\hline
2019 & 72 & 72 & 72 & 96 & 48 & 72 & 24 & 216 \\
2020 & 72 & 72 & 68 & 72 & 36 & 104 & 0 & 212 \\
2021 & 76 & 76 & 76 & 72 & 48 & 108 & 0 & 228 \\
2022 & 72 & 72 & 66 & 70 & 44 & 96 & 22 & 210 \\
2023 & 68 & 68 & 64 & 24 & 84 & 92 & 24 & 200 \\
2024 & 68 & 68 & 68 & 36 & 72 & 96 & 24 & 204 \\
2025 & 64 & 64 & 62 & 42 & 64 & 84 & 18 & 190 \\
\hline
\textbf{Total} & \textbf{492} & \textbf{492} & \textbf{476} & \textbf{412} & \textbf{396} & \textbf{652} & \textbf{112} & \textbf{1460} \\
\hline
\end{tabular}%
}
\caption{Year-wise distribution of questions by subject and question type. \textbf{Num.:} Numerical-type questions, \textbf{Match.:} Matching-type questions.}
\label{tab:yearwise-subject-questiontype}
\end{table}

Question type analysis shows that Numerical questions constitute the largest category with 652 instances (44.7\%), followed by MCQ-Multiple with 412 questions (28.2\%) and MCQ-Single with 396 questions (27.1\%). This distribution aligns with the JEE Advanced examination pattern, where numerical answer questions play a prominent role in assessing problem-solving capabilities. 

Additionally, 442 questions (30.3\%) mandate visual interpretation of mathematical diagrams, chemical structures, or physical illustrations, reflecting the multimodal nature of our task (termed "Image Required"). For the remainder, images are optional, albeit still fed in as screenshots to test the models' OCR capabilities (termed "Image Optional").

\section{Results}

We evaluate 17 state-of-the-art VLMs on \jee: sub-10B models (Aya Vision 8B, Qwen 2.5 VL 7B~\cite{bai2025qwen25vltechnicalreport}, InternVL3 8B~\cite{zhu2025internvl3exploringadvancedtraining}), sub-50B models (Kimi VL Thinking 2506 16B, InternVL3.5 14B/30B, Gemma3 27B~\cite{gemmateam2025gemma3technicalreport}), sub-200B models (InternVL3 78B, Llama4 Scout 109B), and large open models (Qwen3 VL 235B, Llama4 Maverick 400B). For closed-source models, we test OpenAI o3\footnote{\url{https://openai.com/index/introducing-o3-and-o4-mini}}, Gemini 2.5 Pro/Flash, GPT-5/GPT-5-mini, and Grok 4 Fast (July-October 2025 API access). All experiments used Nvidia RTX 5080/5090 GPUs. Prompts are detailed in Appendix~\ref{sec:prompts-used}.

\subsection{Performance analysis of contemporary VLMs}
\label{results:acc}




Table~\ref{tab:big_evals_table} presents the Pass@1 accuracy averaged over $k=10$ runs (Appendix~\ref{sec:k-value-appendix} justifies this choice) of all evaluated models across subjects and question types (RQ1). Our motivation for selecting Pass@1 is two-fold: (a) OpenAI Simple Evals\footnote{\url{https://github.com/openai/simple-evals}} promotes it, and (b) allows fair comparison with other industry standard vision benchmarks. 

Open-source models reach 9-51\% on the 2025 held-out set while frontier models achieve 73-80\%, a gap of 28-37 percentage points. This frontier advantage holds across all benchmarks tested, but mmJEE-Eval avoids the ceiling effects that compress differences elsewhere: Qwen3 VL scores 85\% on MathVista but only 45\% on mmJEE-Eval.

\begin{table*}[h]
\centering
\footnotesize
\setlength{\tabcolsep}{4pt}
\renewcommand{\arraystretch}{1.2}
\resizebox{\textwidth}{!}{%
\begin{tabular}{lcccccccc}
\toprule
\textbf{Model} & 
\multicolumn{4}{c}{\textbf{mmJEE-Eval (Ours)}} & 
\multicolumn{4}{c}{\textbf{Other Industry Standard Benchmarks}} \\
\cmidrule(lr){2-5} \cmidrule(lr){6-9}
 & 
\makecell{\textbf{Acc. \%}\\\textbf{(Full)}} & 
\makecell{\textbf{Acc. \%}\\\textbf{(2025 set)}} & 
\makecell{\textbf{Marks}\\\textbf{(\%)}} & 
\makecell{\textbf{Marks w/}\\\textbf{CT (\%)}} &
\makecell{\textbf{MMMU}\\\textbf{}} &
\makecell{\textbf{MMMU}\\\textbf{Pro}} &
\makecell{\textbf{Math}\\\textbf{Vista}} &
\makecell{\textbf{Char}\\\textbf{Xiv}} \\
\midrule
Random choice & 8.3 ±0.3 & 9.3 ± 1.6 & -46 (-12.7\%) & -7 (-1.9\%) & 22.1\% & 12.6\% & — & 10.8\% \\

\hline

Aya Vision 8B & 8.5 ±0.6 & 9.2 ± 1.0 & -45 (-12.5\%) & -16 (-4.4\%) & 39.9\% & — & — & — \\
Kimi VL Thinking 2506 16B & 9.8 ± 0.2 & 11.2 ± 2.7 & -16 (-4.44\%) & -11 (-3.06\%) & 64.0\% & 46.3\% & 68.7\% & — \\
Qwen 2.5 VL 7B & 10.9 ±0.8 & 12.4 ± 1.9 & -21 (-5.83\%) & 5 (1.4\%) & 58.6\% & 46.2\% & 68.2\% & 42.5\% \\
InternVL3 8B & 10.8 ±0.5 & 12.5 ± 1.3 & -22 (-6.1\%) & 13 (3.6\%) & 62.7\% & — & 71.6\% & 37.6\% \\
InternVL 3.5 14B & 15.6 ±0.9 & 16.1 ± 1.6 & 18 (5.0\%) & 26 (7.2\%) & 73.3\% & — & 80.5\% & — \\

InternVL 3.5 30B & 18.8 ±1.2 & 18.8 ± 1.1 & 22 (6.1\%) & 26 (7.2\%) & 75.6\% & — & 80.9\% & — \\
Grok 4 Fast & 19.3 ±1.1 & 17.6 ± 1.5 & 27 (7.5\%) & 40 (11.1\%) & — & — & — & — \\
InternVL3 78B & 22.0 ±0.8 & 23.6 ± 3.4 & 30 (8.33\%) & 44 (12.22\%) & 72.2\% & — & 79.0\% & 46.0\% \\

\hline

Human (qualifying cutoff) & — & — & — & 74 (20.6\%) & 76.2\% & 73\% & — & — \\

\hline

Gemma 3 27B & 29.8 ±0.3 & 30.6 ± 0.7 & 67 (18.6\%) & 79 (21.9\%) & 64.9\% & — & 63.3\% & — \\
Llama4 Scout 109B & 40.4 ±0.5 & 37.1 ± 2.3 & 101 (28.1\%) & 122 (33.9\%) & 69.4\% & — & 70.7\% & — \\
Qwen3 VL 235B Instruct & 44.6 ± 0.4 & 46.2 ± 4.9 & 130 (36.1\%) & 134 (37.2\%) & 78.7\% & 68.1\% & 84.9\% & — \\
Llama4 Maverick 400B & 50.9 ±0.3 & 43.8 ± 2.8 & 118 (32.8\%) & 129 (35.8\%) & 73.4\% & — & 73.7\% & — \\
OpenAI o3 & 77.4 ±0.7 & 72.7 ± 3.7 & 240 (66.7\%) & 241 (66.9\%) & 82.9\% & 76.4\% & 81.1\% & 78.6\% \\
GPT-5-mini ‘Medium’ & 75.3 ± 0.7 & 71.3 ± 6.7 & 255 (70.8\%) & 259 (71.94\%) & 80.0\% & — & — & — \\
Gemini 2.5 Pro & 81.2 ±0.7 & 77.4 ± 0.8 & 254 (70.6\%) & 285 (79.2\%) & 84.0\% & 71.2\% & 84.6\% & — \\
Gemini 2.5 Flash 09-2025 & 82.6 ± 0.4 & 79.8 ± 1.2 & 300 (83.3\%) & 300 (83.3\%) & 79.7\% & — & 81.2\% & — \\ 
GPT-5 & 83.9 ±0.6 & 79.5 ± 1.5 & 291 (80.8\%) & 295 (81.9\%) & 84.2\% & 78.4\% & 82.7\% & 81.1\% \\

\hline

Human (Top 10 performers) & — & — & — & 324 (90.1\%) & 82.6\% & 80.8\% & — & — \\
Human (Rank 1, Topper) & — & — & — & 332 (92.2\%) & 88.6\% & 85.4\% & 60.3\% & 80.5\% \\

\hline

\makecell[l]{$\Delta$ over human\\(Human – Best model)} & — & — & — & 10.1\% & 4.4\% & 7\% & –24.3\% & –0.6\% \\
\bottomrule
\end{tabular}%
}
\caption{Comparison of model performance on mmJEE-Eval (ours) and other benchmarks, MMMU, MMMU Pro, Math Vista, and CharXiv. For mmJEE-Eval, \textbf{Acc. \% (Full)} represents Pass@1 accuracy on the full set of 1,460 questions, \textbf{Acc. \% (2025 set)} presents Pass@1 accuracy on the held-out 2025 subset, \textbf{Marks (\%)} represents total score on the two papers of JEE Advanced 2025 following the official marking scheme, \textbf{Marks w/ CT (\%)} presents confidence thresholded scores. For other benchmarks, we source Pass@1 accuracies from respective leaderboards.}
\label{tab:big_evals_table}
\end{table*}

\begin{table*}[h]
\centering
\footnotesize
\begin{tabular}{lcccccccc}
\hline
\multirow{2}{*}{\textbf{Model}} 
& \multirow{2}{*}{\textbf{Pass@1}} 
& \multirow{2}{*}{\textbf{EP(\%)}} 
& \multirow{2}{*}{\textbf{EC(\%)}} 
& \multicolumn{2}{c}{\textbf{EP→EC (\%)}} 
& \multirow{2}{*}{\begin{tabular}{c}\textbf{Pass@1}\\\textbf{w/ EP/EC}\end{tabular}}
& \multirow{2}{*}{\textbf{Pass@3}} 
& \multirow{2}{*}{\textbf{Pass@5}} \\
\cline{5-6}
& & & & \textbf{EP} & \textbf{EC} & & & \\
\hline
\textbf{Qwen2.5 VL 7B} & 10.6\% & 66.10 & 1.86 & 73.23 & 1.06 & 12.2\% & 41.2\% & 52.9\% \\

\textbf{InternVL3.5 30B} & 18.8\% & 79.50 & 4.96 & 40.55 & 4.10 & 21.9\% & 44.1\% & 64.7\% \\

\textbf{InternVL3 78B} & 22.2\% & 30.30 & 8.51 & 28.87 & 1.86 & 25.8\% & 44.1\% & 67.6\% \\

\textbf{Gemma3 27B} & 29.8\% & 47.16 & 6.45 & 47.08 & 5.05 & 34.2\% & 64.7\% & 85.3\% \\

\textbf{Llama4 Scout} & 40.4\% & 72.57 & 5.13 & 72.91 & 3.47 & 43.5\% & 79.4\% & 88.2\% \\

\textbf{GPT-5} & 83.9\% & 21.90 & 20.77 & 19.89 & 5.20 & 86.3\% & 100.0\% & 100.0\% \\
\hline
\end{tabular}
\caption{Performance comparison of various models across Error Presence (EP) and Error Correction (EC) steps. In EP→EC, we chain the two steps in the same chat instance, triggering EC only if the model successfully detects an error in its previous erroneous instance. We note that performance improvements with EP/EC remain significantly behind those with Pass@k, hinting at severe metacognition gaps in contemporary VLMs.}
\label{tab:ep_ec_passk_combined}
\end{table*}

\textbf{Confidence Thresholding.} Following \citet{arora-etal-2023-llms}, we apply confidence-based selective answering to JEE Advanced 2025's 190 MCQs. For single-correct and multiple-correct MCQs, we implement self-consistency~\cite{wang2022self} and generate responses in 10 attempts to measure consistency. Appendix~\ref{sec:conf-thresholding} details ideal threshold measurement. Table~\ref{tab:big_evals_table} shows that weak models benefit dramatically (Qwen 2.5 VL: +117\%, InternVL3 8B: +157\%), while strong models improve modestly (Gemini 2.5 Pro: +12\%, Flash: +0\%).

\textbf{Human baselines.} Since confidence thresholding to avoid negative marking is part of the human model for JEE Advanced, and only total scores have been released for the exam, we report human baselines in the \textbf{Marks w/ CT (\%)} column of Table~\ref{tab:big_evals_table}. Open models and Grok 4 Fast score below the 50th percentile of human test-takers, with most failing the mmJEE-Eval test (scores below qualifying cutoffs). In contrast, we notice a pattern in closed-frontier models from Google and OpenAI: Gemini Flash/Pro, o3, GPT-5/Mini exceed the 90th percentile. Notably, Gemini 2.5 Flash achieves 300/360 marks (83.3\%) despite being a light model, placing 32 points below the human topper (332/360, 92.2\%).

\subsection{Metacognitive Self-Correction}
\label{results:meta}

We evaluate whether VLMs can detect and correct their reasoning errors (RQ3) using error presence detection (EP), and error correction (EC) following~\citet{li-etal-2024-evaluating-mathematical}. This accurately captures how humans behave in a realistic exam setting, where students first come up with an initial solution, then detect errors, and finally fix them. Table~\ref{tab:ep_ec_passk_combined} compares EP/EC against pass@k sampling. Models, both open and frontier, show 2.2-6.7\% improvement in total score, much lower than the 30.4\% improvement with Pass@3 on average.

Most strikingly, on average, models detect errors in 52.9\% of cases but correct only 8\% of those detected. The EP→EC conversion rate averages 3.5\% across open models. GPT-5 exhibits an anomaly where it scores only 21.9\% in EP (worst result) but 20.8\% in EC (best result) simultaneously. However, when chained together, overall correction capabilities degrade (5.2\% overall EP→EC accuracy). For some context, the model logged a 14.6\% pass@3 improvement. This suggests that while models can eventually get questions correct with sufficient retries, they fail to do so when presented with previous erroneous generations. This is a metacognitive limitation in current-generation models, regardless of size, that mmJEE-Eval uniquely tests compared to standard VLM reasoning benchmarks.

\subsection{Cross-Lingual Consistency}
\label{results:cross_lingual}

\begin{figure}[h]
    \centering
    \includegraphics[width=\columnwidth]{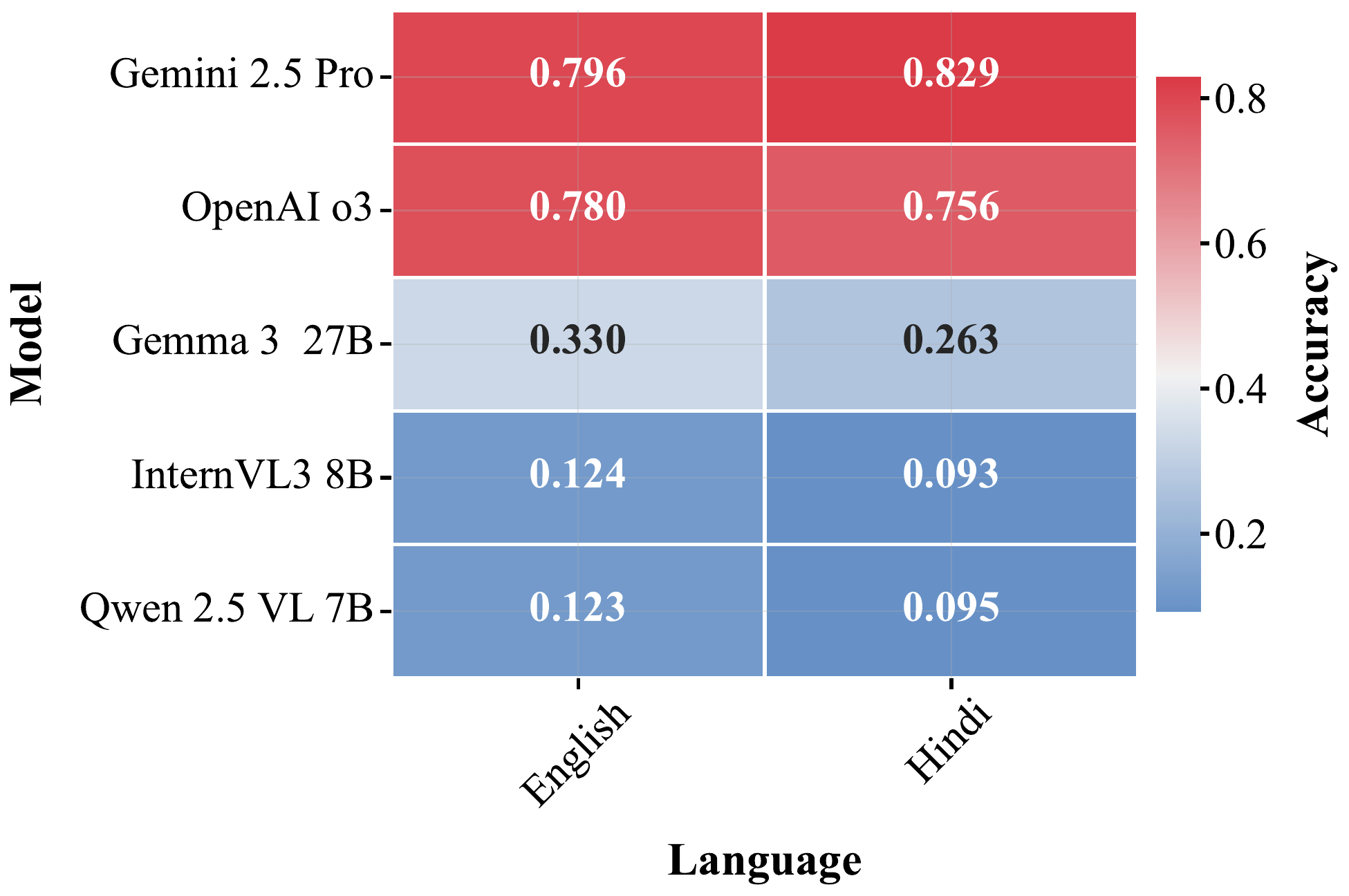}
    \caption{Heatmap comparing accuracy across languages (English, Hindi) for four language models. Gemini 2.5 Pro achieves consistently high performance (>0.80), while smaller models show uniformly low performance (~0.10). Color scale represents accuracy from 0.1 (light blue) to 0.8+ (dark red).}
    \label{fig:multimodal_comp}
\end{figure}

Table~\ref{tab:model_language_comparison} shows model consistency on English-Hindi question pairs from JEE Advanced 2025 by evaluating model responses to identical questions in both Hindi and English, and Figure~\ref{fig:multimodal_comp} shows how average performance varies by language (RQ2). We categorize outcomes into four cases: (a) both Hindi and English responses are correct, (b) both are incorrect, (c) only the English response is correct, and (d) only the Hindi response is correct. Frontier models maintain both languages correctly 67-79\% of the time, while OSS models achieve only 2-36\%.  


We also note higher language-specific failure (Eng+/Hin- or Eng-/Hin+) rates on open models, hinting at English-dominant training (Llama4 models show 17-19\% Eng+/Hin- but only 9-10\% Eng-/Hin+). This asymmetry is more subtle in frontier models (GPT-5: 7.0\% vs 3.6\%, o3: 9.99\% vs 8.37\%). Interestingly, Gemini 2.5 Pro scores higher on Eng-/Hin+ (10.64\%) than Eng+/Hin- (6.93\%), reversing the trend, while Flash maintains the typical English advantage (7.10\% vs 5.47\%). We attribute this Pro-specific gain to a mix of larger model size and Gemini 2.5's 400-language pretraining, where specific gains have been internally noted in Indic languages by the Gemini team~\cite{comanici2025gemini25pushingfrontier}.


\begin{table*}[h]
\footnotesize
\centering
\begin{tabular}{lcccc}
\hline
\textbf{Model Name} & \textbf{Both Correct (\%)} & \textbf{Both Incorrect (\%)} & \textbf{Eng+, Hin- (\%)} & \textbf{Eng-, Hin+ (\%)} \\
\hline
\textbf{Random Baseline} & 1.67 ± 0.36 & 84.89 ± 0.89 & 6.58 ± 0.53 & 6.86 ± 0.66 \\
\hline
\textbf{Qwen 2.5 VL 7B} & 2.87 ± 0.51 & 81.14 ± 1.34 & 9.35 ± 0.69 & 6.64 ± 1.03 \\
\textbf{InternVL 3.5 30B} & 4.41 ± 0.79 & 66.80 ± 2.22 & 22.81 ± 1.84 & 5.98 ± 1.36 \\
\textbf{InternVL 3.5 14B} & 4.80 ± 1.01 & 73.69 ± 1.55 & 15.20 ± 1.71 & 6.31 ± 0.53 \\
\textbf{Grok 4 Fast} & 5.58 ± 0.60 & 67.53 ± 1.43 & 21.07 ± 0.88 & 5.81 ± 0.53 \\
\textbf{InternVL3 78B} & 7.22 ± 1.33 & 62.78 ± 1.37 & 21.98 ± 1.34 & 8.02 ± 1.32 \\
\textbf{Gemma 3 27B} & 16.42 ± 0.28 & 56.67 ± 0.64 & 16.85 ± 0.77 & 10.07 ± 0.65 \\
\textbf{Llama4 Scout 109B} & 27.31 ± 0.98 & 46.48 ± 0.62 & 17.03 ± 1.26 & 9.18 ± 0.82 \\
\textbf{Llama4 Maverick 400B} & 36.26 ± 0.24 & 34.25 ± 0.46 & 19.47 ± 0.96 & 10.02 ± 0.83 \\
\textbf{OpenAI o3} & 72.24 ± 0.47 & 12.22 ± 0.64 & 9.14 ± 0.47 & 6.39 ± 0.36 \\
\textbf{Gemini 2.5 Flash} & 75.96 ± 0.91 & 11.34 ± 0.45 & 7.00 ± 0.89 & 5.70 ± 0.63 \\
\textbf{Gemini 2.5 Pro} & 72.60 ± 1.83 & 9.91 ± 0.87 & 6.76 ± 1.10 & 10.73 ± 0.95 \\
\textbf{GPT-5} & 79.60 ± 1.16 & 9.44 ± 2.37 & 7.25 ± 1.90 & 3.71 ± 0.46 \\
\hline
\end{tabular}
\caption{Model performance comparison across languages in the 2025 edition of JEE Advanced. 
\textbf{Eng+, Hin-}: English correct but Hindi incorrect; 
\textbf{Eng-, Hin+}: English incorrect but Hindi correct.}
\label{tab:model_language_comparison}
\end{table*}

\subsection{Source of error analysis}
\label{subsec:error_analysis}




To understand which factors contribute to mmJEE-Eval's higher difficulty (RQ4), we review 400 model outputs through a hybrid LLM-as-a-Judge (LaaJ) and human-in-the-loop verification pipeline across five representative models using a Bloom's taxonomy-inspired\footnote{\url{https://teaching.cornell.edu/resource/blooms-taxonomy}} framework~\cite{ijcai2024p381,arora-etal-2023-llms}. LaaJ was implemented as an additional evaluator with Claude 4.5 Sonnet to avoid overlaps with the tested models. It first rates responses on four dimensions (0-10 scale): conceptual understanding, grounding in problem context, computational accuracy, and instruction following, which were designed in accordance with the first four Bloom's taxonomy levels (Remember, Understand, Apply, and Analyze). Finally, the authors review the assigned scores and reasoning chains before accepting or rejecting an annotation. 

Table~\ref{tab:human_eval_quality} shows Llama4 Scout 109B (40.4\% accuracy) and GPT-5 (83.9\% accuracy) score nearly identically in overall reasoning quality (7.07 vs 7.08). Scout matches or exceeds GPT-5 on conceptual errors (7.43 vs 6.9), grounding (7.71 vs 6.8), and computation (7.29 vs 6.9), but falls substantially behind on instruction following (5.86 vs 7.7). We conclude that accuracy gaps don't directly correlate with better reasoning quality.

As an illustration, Figure~\ref{fig:model_response_example} shows a representative error from OpenAI o3, where the model demonstrates correct mathematical reasoning but incorrectly assumes "uniform thickness" for wedge-shaped glass pieces despite the figure showing otherwise. This leads to an incorrect answer despite correct subsequent steps, exemplifying the challenge put forward by \jee~(RQ3).


\begin{table*}[h]
\centering
\footnotesize
\begin{tabular}{lccccc}
\hline
\textbf{Model} & \textbf{Conceptual} & \textbf{Grounding} & \textbf{Computation} & \textbf{Instruction} & \textbf{Average} \\
 & \textbf{Errors} & \textbf{Errors} & \textbf{Errors} & \textbf{Errors} & \textbf{} \\
\hline
Grok 4 Fast & 4.2 $\pm$ 1.92 & 4.0 $\pm$ 1.73 & 3.8 $\pm$ 1.79 & 5.4 $\pm$ 1.52 & 4.35 \\

InternVL3 78B & 5.2 $\pm$ 0.84 & 4.6 $\pm$ 1.14 & 4.8 $\pm$ 1.1 & 7.80 $\pm$ 0.45 & 5.6 \\

Llama4 Maverick 400B & 5.83 $\pm$ 1.33 & 5.83 $\pm$ 1.83 & 6.17 $\pm$ 2.14 & 6.0 $\pm$ 1.55 & 5.96 \\

Llama4 Scout 109B & 7.43 $\pm$ 1.9 & 7.71 $\pm$ 1.8 & 7.29 $\pm$ 2.14 & 5.86 $\pm$ 1.07 & 7.07 \\

GPT-5 & 6.9 $\pm$ 2.38 & 6.8 $\pm$ 2.44 & 6.9 $\pm$ 2.42 & 7.7 $\pm$ 1.16 & 7.08 \\
\hline
\end{tabular}
\caption{Source of error analysis results of our hybrid LaaJ and human-in-the-loop study on 400 model outputs of a few representative models. All values are reported on a scale of 0-10, with higher being better.}
\label{tab:human_eval_quality}
\end{table*}

\subsection{Dataset contamination tests}
\label{results:contamination}

Table~\ref{tab:contamination_test} compares performance on 2019-2024 questions (potentially seen during training) versus the 2025 held-out set (exam conducted May 18, 2025; answers released June 2, 2025). Open models show stable or slightly improved performance on 2025 (+0.2\% to +1.9\%), likely reflecting sample variance given N=190. Frontier models drop 2-5\%: GPT-5 (-5.3\%), Gemini 2.5 Pro (-4.3\%), OpenAI o3 (-4.7\%), keeping training set overlap for 2019-2024 public exam papers modest.

Critically, models released after June 2, 2025 (Qwen3 VL, Grok 4 Fast) show no advantage over earlier models, meaning answer keys did not leak into training. The 28-37 point frontier-open gap persists on 2025 data, further proving that mmJEE-Eval is not a memorization artifact. With every year, fresh JEE Advanced questions can be augmented to mmJEE-Eval following our methodology (Section~\ref{sec:methodology}), limiting contamination concerns.

\begin{table*}[h]
\footnotesize
\centering
\begin{tabular}{lccccc}
\hline
\textbf{Model} & \textbf{Release} & \textbf{Knowledge} & \textbf{2019-2024} & \textbf{2025} & \textbf{$\Delta$} \\
 & \textbf{Date} & \textbf{Cutoff Date} & \textbf{Acc (\%)} & \textbf{Acc (\%)} &  \\
\hline
Random baseline & --- & --- & $8.20 \pm 0.52$ & $9.26 \pm 1.56$ & +1.06\% \\
\hline
Aya Vision 8B & Mar 2025 & Late 2024 & $8.07 \pm 2.50$ & $9.21 \pm 10.03$ & +1.14\% \\

Kimi VL A3B Thinking 2506 16B & Jun 2025 & Unknown & $9.88 \pm 0.50$ & $11.05 \pm 13.37$ & +1.17\% \\

Qwen 2.5 VL 7B & Jan 2025 & Nov 2024 & $10.64 \pm 0.83$ & $12.37 \pm 1.96$ & +1.73\% \\

InternVL3 8B & Apr 2025 & Aug 2024 & $10.64 \pm 0.42$ & $12.54 \pm 1.28$ & +1.91\% \\

InternVL3.5 14B & Aug 2025 & Aug 2024 & $15.95 \pm 1.11$ & $16.11 \pm 1.64$ & +0.15\% \\

Grok 4 Fast & Sep 2025 & Nov 2024 & $19.32 \pm 0.99$ & $17.62 \pm 1.46$ & -1.70\% \\

InternVL3.5 30B & Aug 2025 & Aug 2024 & $18.90 \pm 0.90$ & $18.84 \pm 1.07$ & -0.06\% \\

InternVL3 78B & Apr 2025 & Aug 2024 & $21.99 \pm 0.61$ & $23.60 \pm 3.43$ & +1.61\% \\

Gemma 3 27B & Mar 2025 & Aug 2024 & $29.48 \pm 0.27$ & $30.62 \pm 0.72$ & +1.14\% \\

Llama4 Scout 109B & Apr 2025 & Aug 2024 & $40.87 \pm 0.63$ & $37.07 \pm 2.28$ & -3.81\% \\

Llama4 Maverick 400B & Apr 2025 & Aug 2024 & $51.97 \pm 0.55$ & $43.83 \pm 2.79$ & -8.13\% \\

Qwen3 VL 235B A22B Instruct & Sep 2025 & Unknown & $44.52 \pm 0.00$ & $45.03 \pm 0.00$ & +0.50\% \\

OpenAI o3 & Apr 2025 & Jun 2024 & $77.41 \pm 0.66$ & $72.70 \pm 3.65$ & -4.71\% \\

Gemini 2.5 Pro & Mar 2025 & Jan 2025 & $81.73 \pm 0.83$ & $77.39 \pm 0.79$ & -4.33\% \\

GPT-5 & Aug 2025 & Sep 2024 & $84.75 \pm 0.63$ & $79.47 \pm 1.49$ & -5.27\% \\

Gemini 2.5 Flash 09-2025 & Sep 2025 & Jan 2025 & $82.43 \pm 0.00$ & $80.53 \pm 0.00$ & -1.90\% \\
\hline
\end{tabular}
\caption{Data contamination tests on mmJEE-Eval. We evaluate effects with (a) model release dates and (b) performance on older (2019-2024 editions) and held-out (2025) JEE Advanced questions. Notably, the 2025 edition was held on May 12, 2025, with answers released on June 2, 2025 (after the cutoff dates of the tested models). The difference between the two sets averages at 2.79\% (range: -8.13\%-1.91\%), keeping any contamination effects modest. Particularly, we notice large drops on o3, Gemini 2.5 Pro, and GPT-5, meaning closed frontiers suffer more from training data memorization than open alternatives, which report gains on the held-out set.}
\label{tab:contamination_test}
\end{table*}

\subsection{Ablations}
\label{results:ablations}

\begin{table}[h]
\centering
\resizebox{1.05\columnwidth}{!}{%
\begin{tabular}{lcccc}
\hline
\textbf{Model} &
\multicolumn{1}{c}{\textbf{Images+}} &
\multicolumn{1}{c}{\textbf{EasyOCR}} &
\multicolumn{1}{c}{\textbf{Gemma3}} &
\multicolumn{1}{c}{\textbf{Diagrams}} \\
& \multicolumn{1}{c}{\textbf{Diagrams}} &
\multicolumn{1}{c}{(text)} &
\multicolumn{1}{c}{\textbf{OCR} (text)} &
\multicolumn{1}{c}{\textbf{Cropped}} \\
\hline
Qwen2.5 VL 7B & 12.37\% & 11.58\% & 11.58\% & 11.58\% \\
Gemma3 27B & 30.62\% & 14.74\% & 33.68\% & 26.32\% \\
InternVL3 78B & 23.60\% & 13.68\% & 28.42\% & 21.58\% \\
Llama4 Scout & 37.07\% & 20.53\% & 31.05\% & 28.42\% \\
Llama4 Maverick & 43.83\% & 19.47\% & 40.53\% & 38.42\% \\
Gemini 2.5 Flash & 79.52\% & 31.05\% & 57.37\% & 66.32\% \\
GPT-5 & 79.47\% & 27.37\% & 61.01\% & 62.63\% \\
\hline
\end{tabular}%
}
\caption{Ablation study showing model performance on the 2025 subset across different setups and OCR pipelines. Reported numbers are mean accuracies.}
\label{tab:ablation_ocr}
\end{table}

First, we isolate visual and textual components by testing four conditions: (1) original images with diagrams, (2) EasyOCR text extraction (known to corrupt complex mathematical notation), (3) Gemma3 27B-based OCR (improved symbolic parsing), and (4) cropped images removing all diagrams. Table~\ref{tab:ablation_ocr} presents results on the 2025 held-out set. 

Frontier models exhibit severe OCR sensitivity: Gemini 2.5 Flash drops 48.5 points with corrupted EasyOCR (79.5\%→31.1\%), recovering 26.3 points with improved Gemma3 OCR (57.4\%), then regaining an additional 8.9 points when diagrams are restored (66.3\%). GPT-5 shows similar degradation. This pattern indicates SoTA models don't simply pattern-match text. Open models exhibit opposite behavior: while EasyOCR corruption drops accuracy, they fully recover with better OCR, with minimal diagram dependence. This suggests open models rely primarily on textual patterns rather than visual-symbolic integration, explaining the ceiling observed in Table~\ref{tab:big_evals_table}.

Even with diagrams removed, Flash achieves 66.3\% while Maverick reaches 40.5\% and Scout 31\%. Since these are all MoEs with \textasciitilde{}17B active parameters, we hypothesize that training methodology supersedes vision capabilities.

\subsection{Discussion}

Among our results, three insights about contemporary VLM evaluation and capabilities stand out. 

First, \textbf{benchmark design matters.} In mmJEE-Eval, we observe a near 50-point performance difference between closed OpenAI and Google frontiers (66.7-83.3\% total marks) and the next-best OSS VLMs (28.1-36.1\% total marks), a separation that does not appear on MathVista or MMMU (RQ1). Through systematic ablations, we attribute this difference to the exam-style evaluation design, multimodal question setup, and bilingual inputs, not just problem complexity (RQ4). Relatedly, OSS models show brittleness to bilingual perturbations, with Llama 4 Maverick is on average 38.84\% behind closed VLMs from OpenAI and Google (RQ2). We also find that exam-style scoring reveals calibration behaviors hidden by standard accuracy metrics: weaker models benefit dramatically from selective answering (+117-157\%), while strong models show calibration (+0-12\%). 

Secondly, we find additional proof that \textbf{training methodology dominates raw scale.} Gemini Flash's 30-40\% better accuracy than Maverick and Scout (all 17B sparse MoEs) indicates superior proprietary training innovations. Further, pass@k accuracy doesn't fully capture reasoning quality gaps: Scout outperforms GPT-5 on Bloom's taxonomy-inspired evaluation despite a 40\% pass@1 gap. This suggests problems in pure breadth-style leaderboard benchmarks, motivating our exam style and error analysis-based evaluation. 

Finally, addressing RQ3, we find \textbf{universal metacognitive limitations in VLMs}. Models can often detect when their reasoning is uncertain (21–73\% of cases) but rarely correct themselves when prompted (1.06–5.2\% improvements), far below simple sampling gains ($\approx$30\% from pass@3). Even GPT-5 corrects only 5.2\% of its errors (vs 14.6\% pass@3 gain). This suggests a problematic finding: while VLMs can eventually arrive at correct solutions, they do not reliably recognize when they are wrong.

Together, these findings position mmJEE-Eval as a strong complementary VLM reasoning benchmark besides breadth-first tests like MMMU.

\section{Conclusion}
We introduced mmJEE-Eval, a multimodal and bilingual benchmark built from 7 years of India’s JEE Advanced examination, designed to evaluate exam-style reasoning in VLMs. Through detailed evaluation, we found a near-50\% performance gap between the leading closed VLMs and the strongest OSS systems. We further find that open models show limited bilingual consistency, often failing to get the same question correct across English and Hindi. Ultimately, we observe VLMs only correct 1.06-5.05\% of their own identified errors. Given typical 30-40\% gains with pass@k sampling, this indicates weak meta-cognition even when reasoning capability is present. 

In essence, mmJEE-Eval presents the first comprehensive tool for auditing reasoning depth, bilingual robustness, and meta-cognitive behavior within a single framework, revealing gaps that other tests do not fully capture.


\section*{Limitations}

Since JEE Advanced is conducted only in English and Hindi, our study is limited to two languages, Hindi and English, in linguistic scope. JEE Advanced comes with additional constraints, such as timing, where all questions must be solved within 3 hours. Our evaluation doesn't model this. Moreover, human test-takers employ iterative problem-solving and error correction capabilities, while our implementation focuses on single-pass EP and EC. 

Additionally, our evaluation does not capture the cultural and educational context that Indian students receive through years of specialized JEE preparation. Finally, while JEE Advanced represents challenging pre-college STEM reasoning, ceiling effects may limit discrimination among frontier models. 

\section*{Acknowledgments}
SPARC (Scheme for Promotion of Academic and Research Collaboration) Phase-III (Project ID: 3385) and TIH IIT Tirupati (IITTNiF/TPD/2024-25/P16) partially supported this research work. We also thank Nvidia India for providing Blackwell-based RTX 50-series GPUs to run the experiments.

\bibliography{custom}

\clearpage

\appendix

\noindent{\LARGE \textbf{Appendices}}
\vspace{0.5cm}

This supplementary material presents additional details on the following aspects:  
\begin{itemize}
    \setlength{\itemsep}{2pt}
    \item \textbf{Appendix A:} Confidence Thresholding Tests
    \item \textbf{Appendix B:} Optimal $k$ Value Selection for Pass@1 Accuracy
    \item \textbf{Appendix C:} Prompts Used
    \item \textbf{Appendix D:} Manual Annotation Software Design 
    \item \textbf{Appendix E:} Studying Failure Cases 
    \item \textbf{Appendix F:} JEE Advanced Scoring System
    \item \textbf{Appendix G:} Datasheet
\end{itemize}

\section{Confidence Thresholding Tests}
\label{sec:conf-thresholding}

\begin{figure*}[h]
    \centering
    \includegraphics[width=\textwidth]{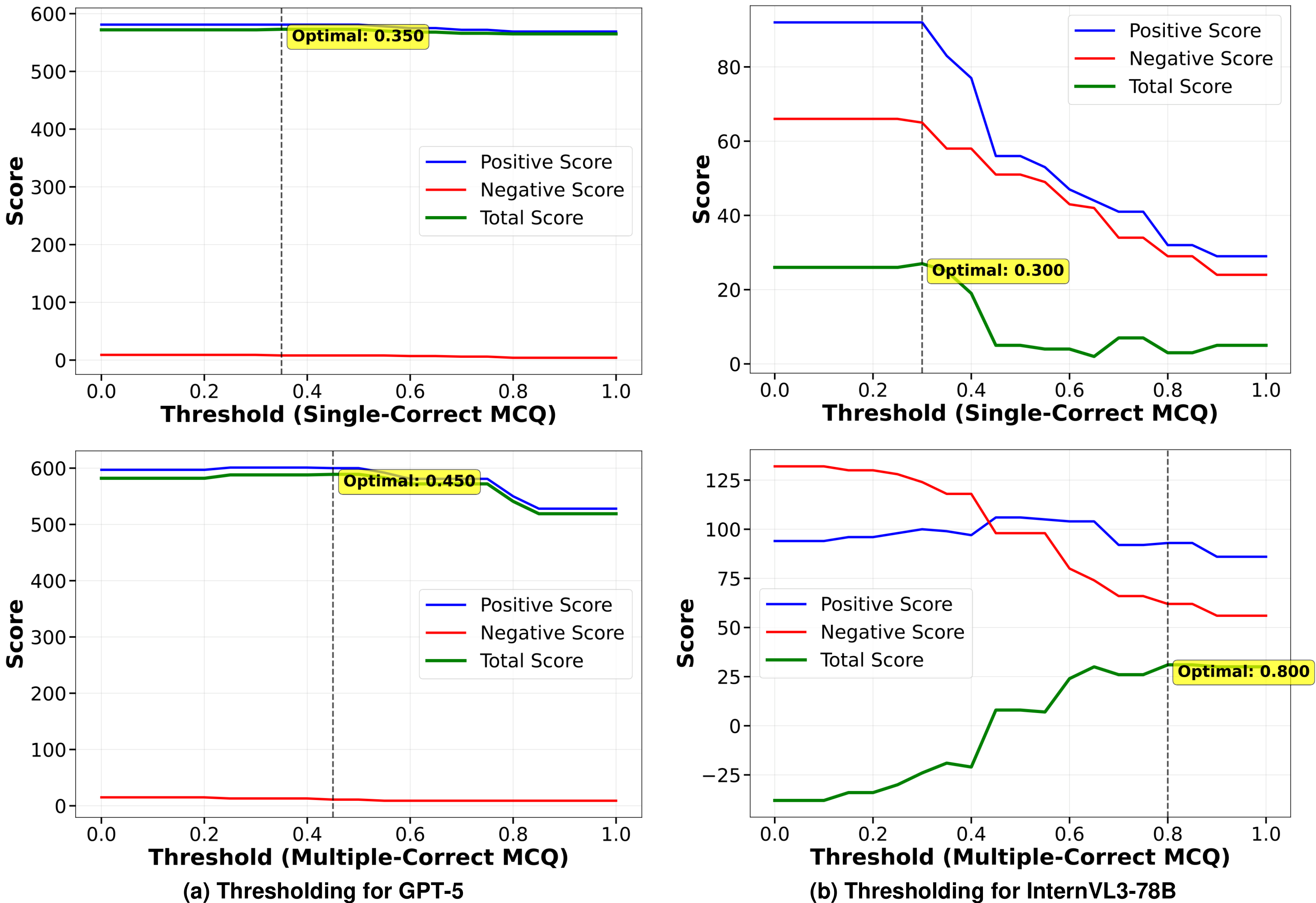}
    \caption{Thresholding analysis comparing GPT-5 and InternVL3-78B performance. (a) shows the thresholding behavior for GPT-5, while (b) demonstrates the corresponding analysis for InternVL3-78B. We note more stable behavior on the OpenAI model than InternVL3, denoting that correct answers were generated more consistently across 10 runs.}
    \label{fig:thresholding_comparison}
\end{figure*}

Figure~\ref{fig:thresholding_comparison} illustrates how positive, negative, and total scores vary with confidence thresholds for GPT-5 and InternVL3-78B. For example, in single-correct MCQ questions, GPT-5's answers are only considered if they appear in at least 35\% of the sampled responses. Similarly, for InternVL 3 78B, the threshold is 30\%. Answers that fail to form a consensus are skipped to avoid negative marking.

Looking at the characteristics of the plots, we note that the closed-source model (GPT-5) maintains relatively stable performance across experimental settings, while InternVL3-78B exhibits erratic behavior with volatile score fluctuations. This provides further evidence for the inconsistent reasoning capabilities and poor calibration of open-source models.

\section{Optimal $k$ Value Selection for Pass@1 Accuracy}
\label{sec:k-value-appendix}

To determine the optimal number of inference runs for reliable pass@1 accuracy estimation, we conducted systematic experiments with Gemma3 27B across varying numbers of runs. This analysis is crucial for balancing computational efficiency with statistical precision in our evaluation methodology.

\begin{figure}
    \centering
    \includegraphics[width=\columnwidth]{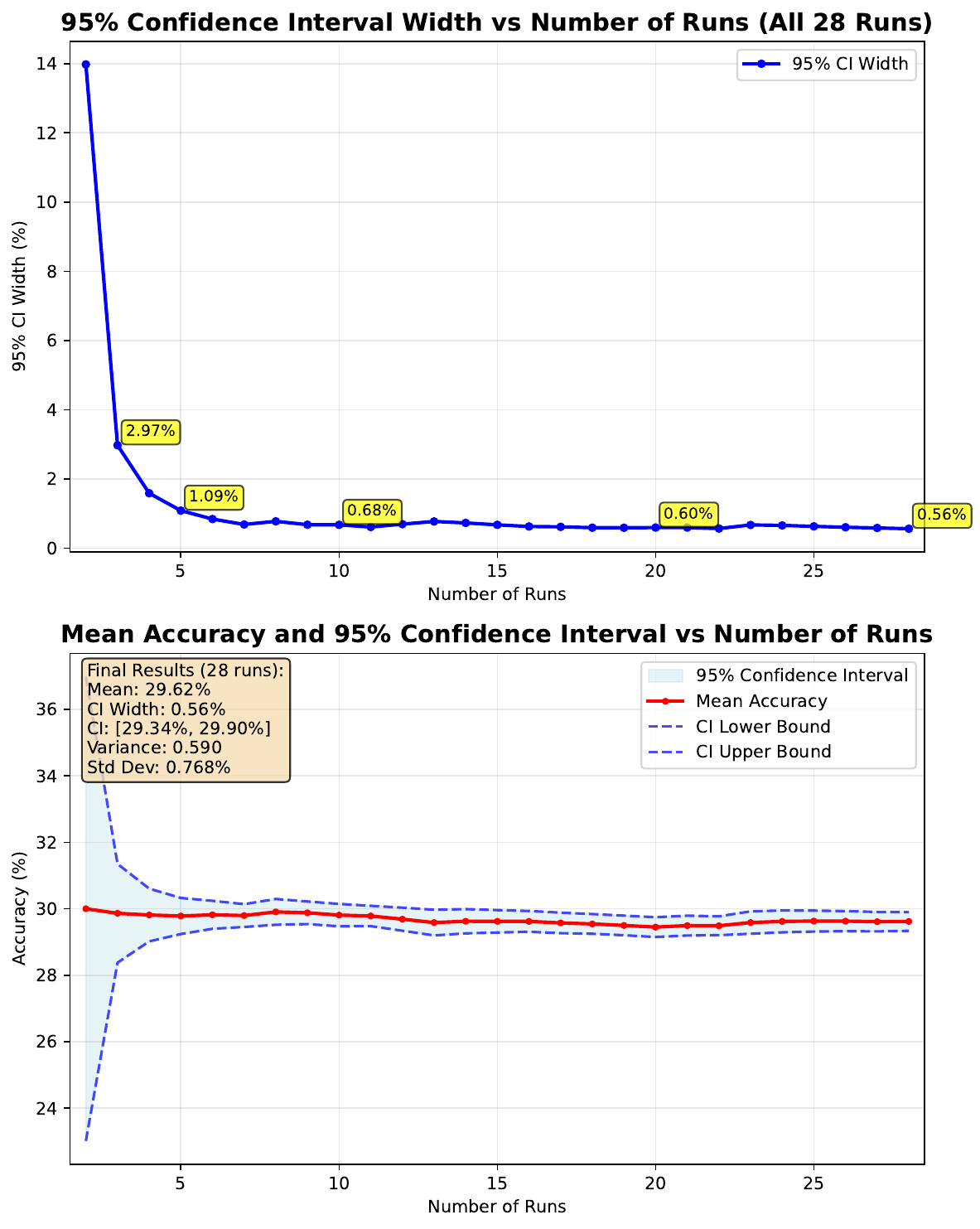}
    \caption{Analysis of confidence interval convergence across multiple experimental runs. The top panel shows how the 95\% confidence interval width decreases as the number of runs increases, demonstrating improved statistical precision. The bottom panel displays the mean accuracy with confidence bounds, showing stabilization of the estimate around 29.62\% after sufficient runs.}
    \label{fig:ci_analysis}
\end{figure}

Figure~\ref{fig:ci_analysis} demonstrates the convergence behavior of confidence intervals as the number of runs increases. The 95\% confidence interval width decreases rapidly from 14.1\% at a single run to 2.97\% at 3 runs, eventually stabilizing at 0.56\% after 28 runs. The mean accuracy stabilizes around 29.62\% after approximately 10 runs, with minimal fluctuation thereafter. This convergence pattern indicates that statistical precision improves substantially with increased sampling until reaching a plateau.

Our analysis reveals distinct accuracy characteristics between different model capability levels. Gemma3 27B exhibits consistent performance with accuracy values clustering around the mean (29.63\%) across 32 runs, showing minimal variance (standard deviation: 0.768\%). The maximum deviation spans only 3.35 percentage points (28.36\% to 31.71\%), indicating stable and reliable measurement characteristics suitable for comparative evaluation.

\begin{figure*}
    \centering
    \includegraphics[width=\textwidth]{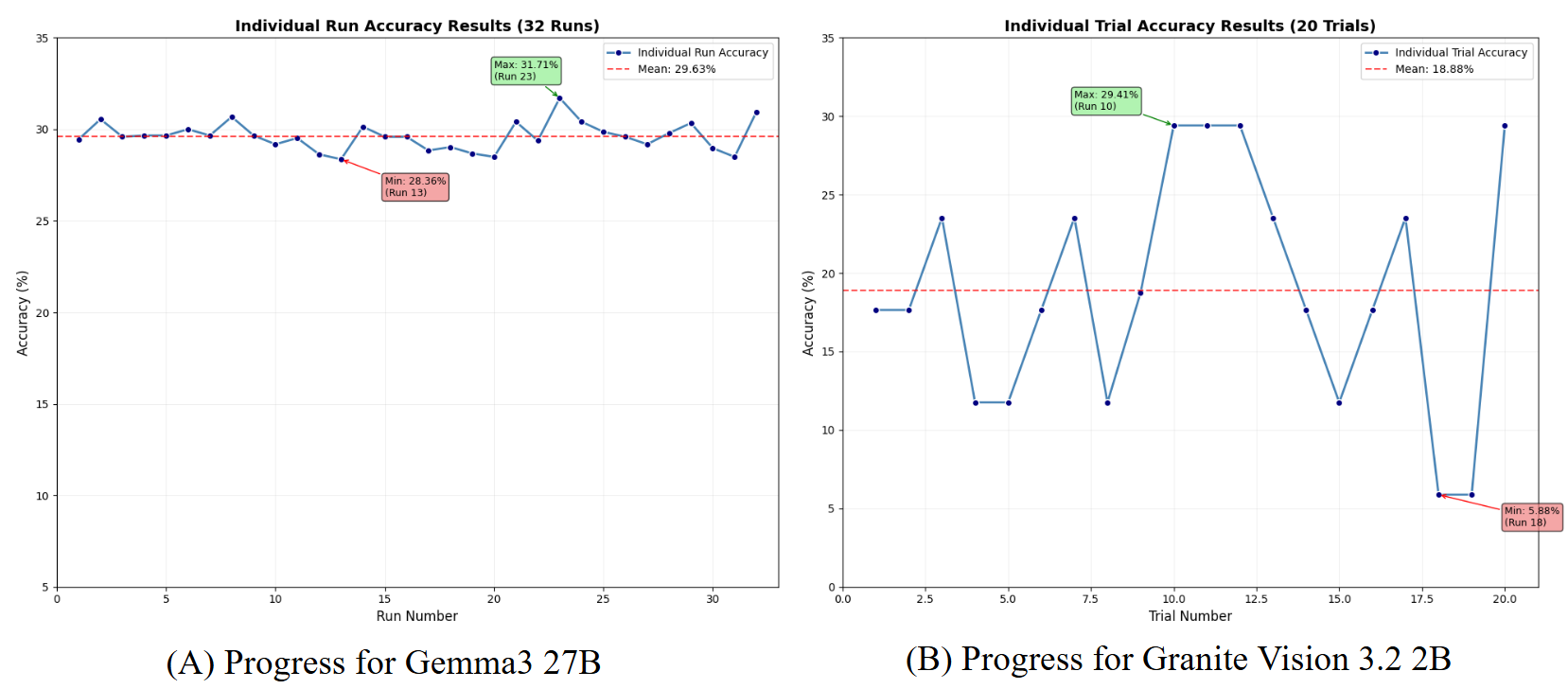}
    \caption{Accuracy characteristics of two open-source models—Gemma3 27B and Granite Vision 3.2 2B—on \jee. Gemma3 27B (left) shows consistent performance with minimal variance around 29.63\% mean accuracy. Granite Vision 3.2 2B (right) exhibits erratic behavior resembling random guessing, with accuracy fluctuations between 5.88\% and 29.41\%, making it unsuitable for reliable evaluation.}
    \label{fig:progress_comp}
\end{figure*}

In contrast, experiments with Granite Vision 3.2 2B (Figure~\ref{fig:progress_comp}) revealed highly erratic accuracy patterns resembling random guessing behavior. Individual trial accuracies fluctuated dramatically between 5.88\% and 29.41\% across 20 trials, with no discernible convergence pattern. This erratic behavior, characteristic of models with insufficient capability for the task complexity, produces unreliable measurements unsuitable for systematic evaluation. Consequently, we exclude such models from our final analysis to maintain evaluation integrity.
Based on this empirical analysis, we selected $k=10$ runs as the optimal balance between computational efficiency and statistical reliability. While confidence intervals stabilize at $k=7$ (Figure~\ref{fig:ci_analysis}), we chose $k=10$ to provide additional headroom for optimization and ensure robust estimates across different model capabilities. This configuration yields a 95\% confidence interval width of approximately 1.09\%, providing sufficient precision for comparative analysis while maintaining practical computational requirements.

\section{Prompts Used}
\label{sec:prompts-used}

We employ different prompting strategies based on the question language and type to ensure optimal model performance across our multilingual dataset. For Hindi questions, we explicitly acknowledge the language barrier while requesting comprehensive analysis, as models often struggle with cross-lingual mathematical reasoning. English questions receive more direct instructions focusing on step-by-step analysis and clear answer presentation.

Our answer extraction methodology uses LaTeX \texttt{\textbackslash\textbackslash boxed\{\}} formatting for all question types, departing from traditional evaluation scripts. This design choice is motivated by extensive evidence in the literature showing that contemporary language models are predominantly trained on LaTeX-formatted mathematical content~\cite{lightman2023lets,hendrycks2021measuring}. Additionally, our preliminary experiments using OpenAI's MATH dataset evaluation script\footnote{\url{https://github.com/hendrycks/math/tree/main/modeling}} yielded significantly lower success rates due to parsing failures and format inconsistencies, confirming that structured LaTeX formatting provides more reliable answer extraction.

Prompts for MCQ-Single and MCQ-Multiple questions emphasize exact option selection with explicit formatting instructions, while Numerical questions require precise decimal formatting within \texttt{\textbackslash\textbackslash boxed\{\}} tags. Matching questions, being the most complex, receive detailed instructions about selecting the single best option that correctly maps items between columns.

\subsection{System Prompts by Language}
\begin{promptbox}[Hindi System Prompt]
\texttt{This is a question image in Hindi. Please analyze the image carefully, understand the question despite the language barrier, reason step-by-step and provide your answer at the end.}
\end{promptbox}

\begin{promptbox}[English System Prompt]
\texttt{Please analyze the image carefully, reason step-by-step and provide your answer at the end.}
\end{promptbox}

\subsection{Question Type-Specific Instructions}
\begin{promptbox}[MCQ-Single Instructions]
\texttt{For this question: - Choose exactly ONE option (A, B, C, or D) - Format your answer in \textbackslash\textbackslash boxed\{\} as just one letter (e.g., \textbackslash\textbackslash boxed\{A\})}
\end{promptbox}

\begin{promptbox}[MCQ-Multiple Instructions]
\texttt{For this question: - Choose ONE OR MORE options (A, B, C, and/or D) - Format your answer in \textbackslash\textbackslash boxed\{\} with letters (e.g., \textbackslash\textbackslash boxed\{ABC\} or \textbackslash\textbackslash boxed\{B\})}
\end{promptbox}

\begin{promptbox}[Numerical Instructions]
\texttt{For this question: - Provide a numerical value - Round to appropriate decimal places if needed - Format your answer in \textbackslash\textbackslash boxed\{\} (e.g., \textbackslash\textbackslash boxed\{2.5\} or \textbackslash\textbackslash boxed\{42\})}
\end{promptbox}

\begin{promptbox}[Matching Instructions]
\texttt{For this question: - Choose ONE option that correctly matches items from left column to right column - Format your answer in \textbackslash\textbackslash boxed{} as just one letter (e.g., \textbackslash\textbackslash boxed\{A\})}
\end{promptbox}

\subsection{Error Presence (EP) Prompts:}

\begin{figure*}[h]
    \centering
    \includegraphics[width=0.9\textwidth]{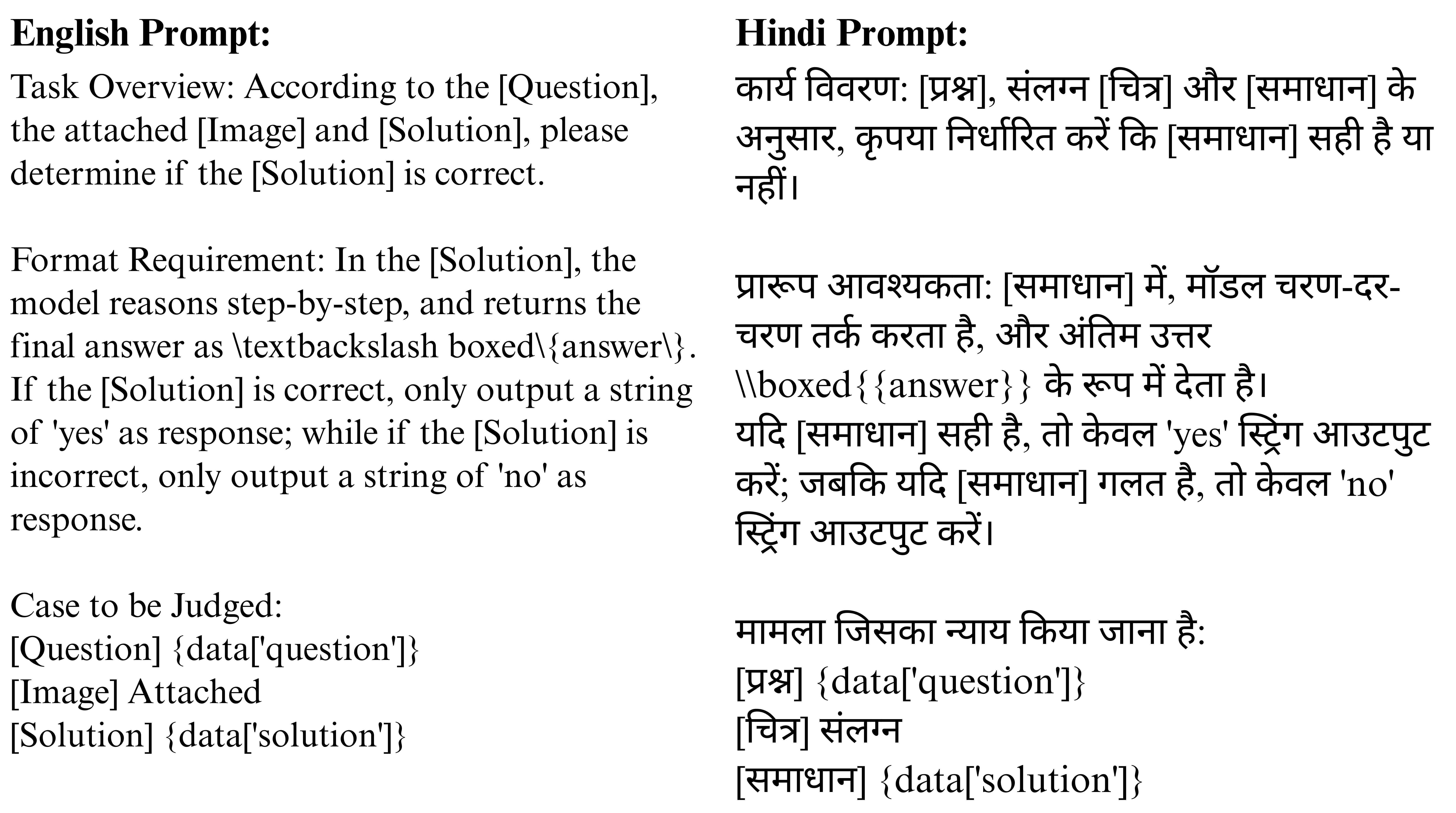}
    \caption{Error Presence (EP) prompt}
    \label{fig:ep_prompt}
\end{figure*}

For error presence detection, we adapted prompts from~\cite{li-etal-2024-evaluating-mathematical} with language-specific modifications as shown in Figure~\ref{fig:ep_prompt}.

\subsection{Error Correction (EC) Prompts:}

\begin{figure*}[h]
    \centering
    \includegraphics[width=0.9\textwidth]{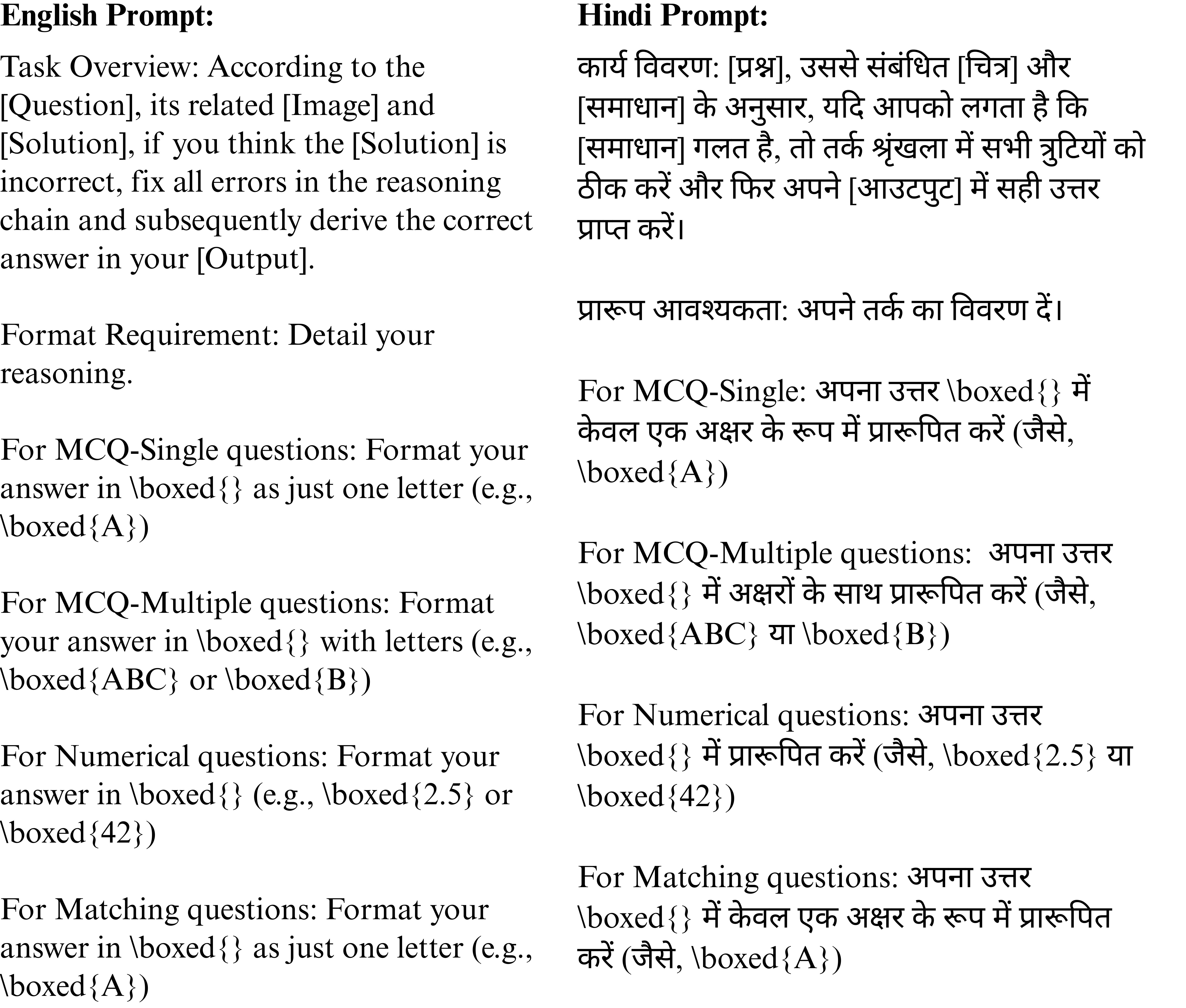}
    \caption{Error Correction (EC) prompt}
    \label{fig:ec_prompt}
\end{figure*}

For error correction, we use the prompt in Figure~\ref{fig:ec_prompt}.

\section{Manual Annotation Software Design}
\label{sec:manual-collection-software}

To ensure accurate and systematic collection of JEE Advanced questions, we developed three custom annotation utilities that streamline the entire data collection and verification pipeline.

\begin{figure}[H]
    \centering
    \includegraphics[width=\columnwidth]{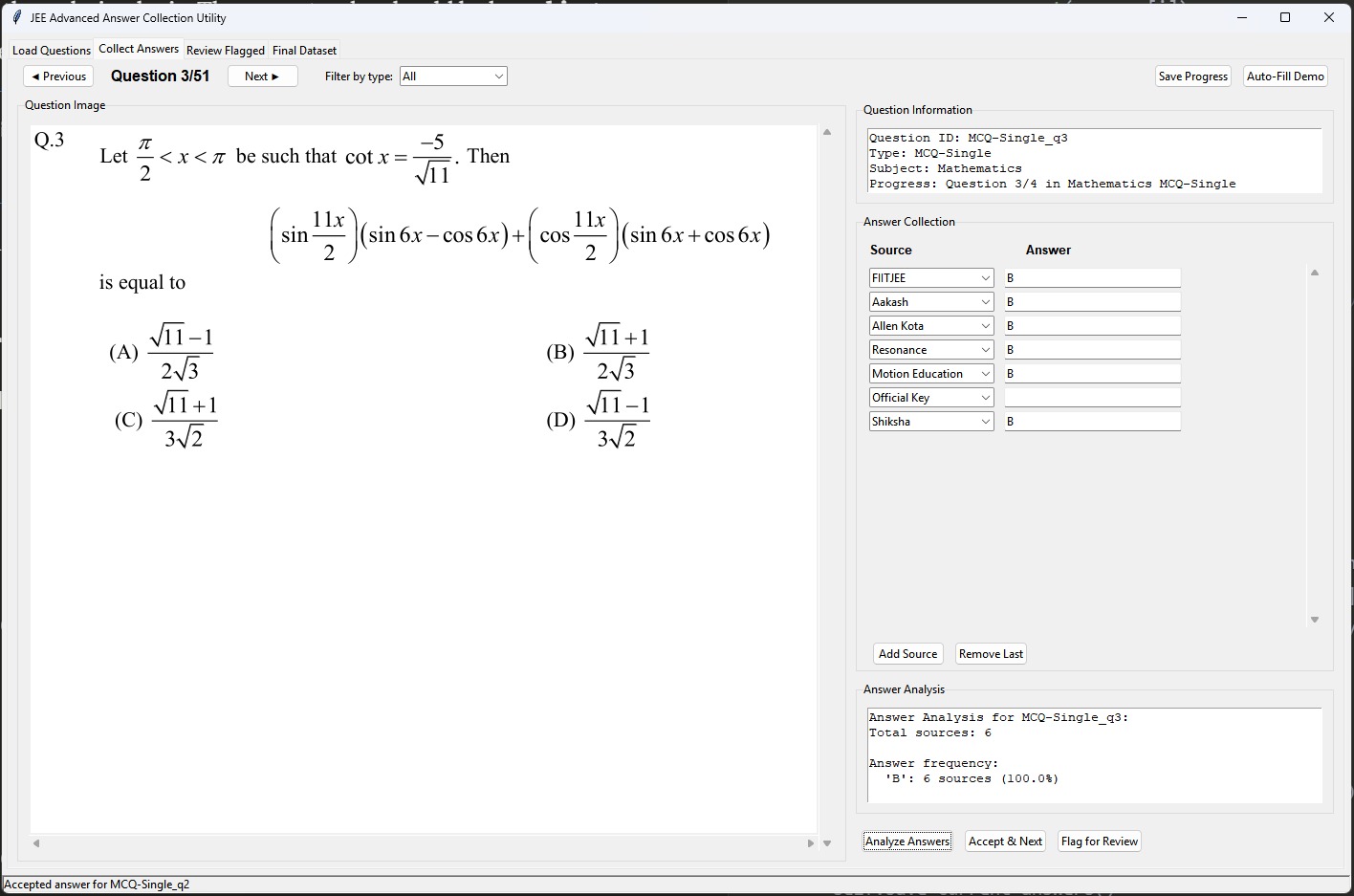}
    \caption{Our custom answer collection utility shows answers collected from five independent coaching institutes (FIITJEE, Aakash, Allen Kota, Resonance, Motion Education), and Official Key. The tool performs automatic majority voting analysis, with answer 'B' received unanimous consensus (100\%) from all 6 sources.}
    \label{fig:answer_collection_tool}
\end{figure}

\begin{figure}[H]
    \centering
    \includegraphics[width=\columnwidth]{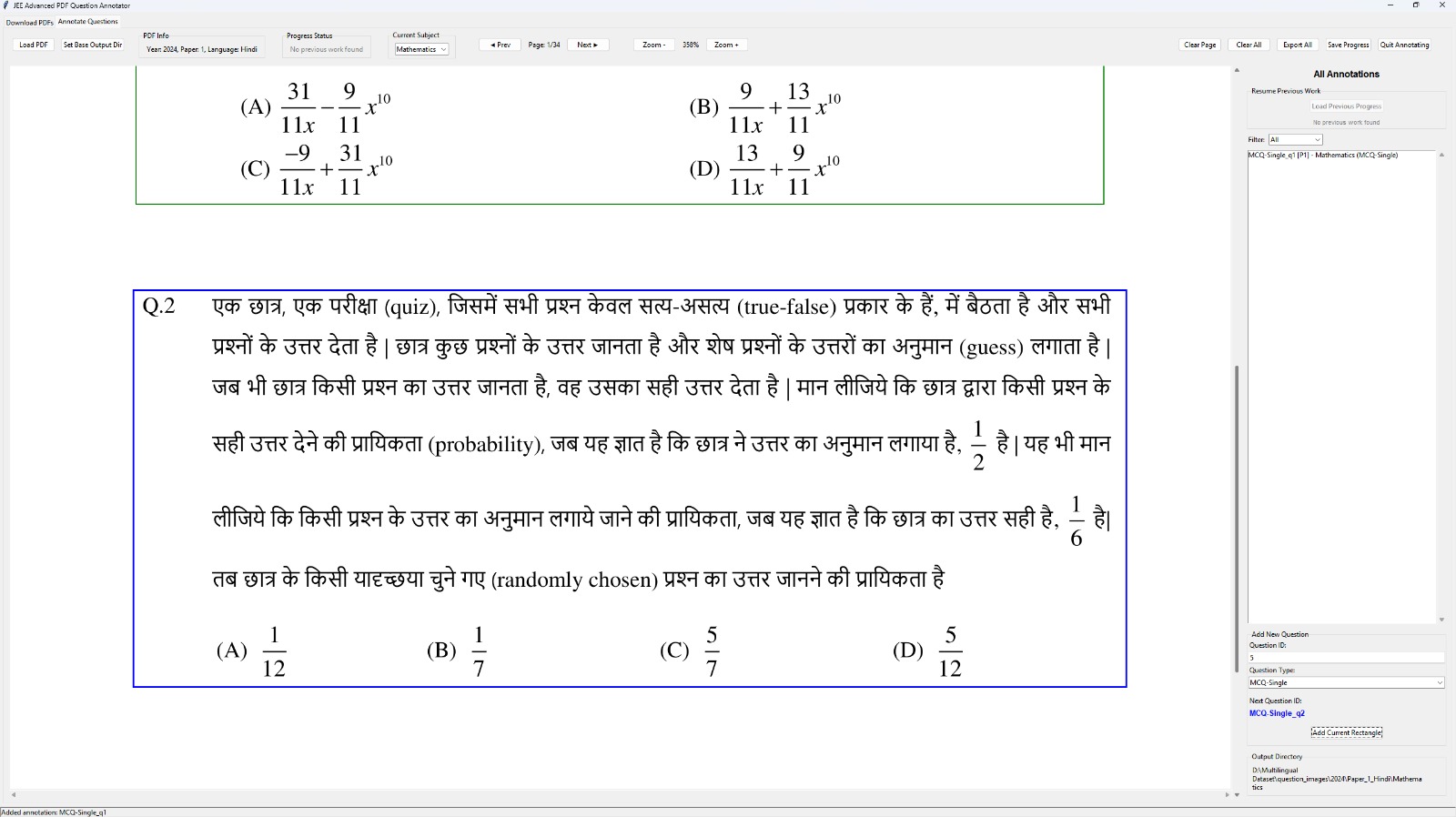}
    \caption{Our custom annotation tool for JEE Advanced question collection automatically downloads question papers and presents them page-by-page to ensure accurate question segmentation and collection.}
    \label{fig:annotation_tool}
\end{figure}

\textbf{Question Annotation Tool.} Our PDF question annotator (Figure~\ref{fig:annotation_tool}) automatically downloads question papers from official sources across all years (2019-2025) and presents them in a page-by-page interface for manual annotation. Annotators can draw precise bounding boxes around individual questions, with support for multi-rectangle questions that span multiple regions or pages. The tool maintains structured metadata including year, paper number, language, subject, and question type, while automatically generating unique question IDs following our naming convention. Key features include zoom controls for detailed annotation, progress tracking across subjects, and the ability to export annotated questions as high-resolution images.

\textbf{Answer Collection Utility.} The answer verification tool (Figure~\ref{fig:answer_collection_tool}) implements our multi-source verification approach by collecting answers from six independent sources: five major coaching institutes (FIITJEE, Aakash, Allen Kota, Resonance, Motion Education) and official answer keys when available. The interface displays each question image alongside a structured form for recording answers from each source. The tool performs automatic majority voting analysis, calculating confidence scores and flagging questions that fail to achieve sufficient consensus (>60\%) for manual review. 

\begin{figure}[H]
    \centering
    \includegraphics[width=\columnwidth]{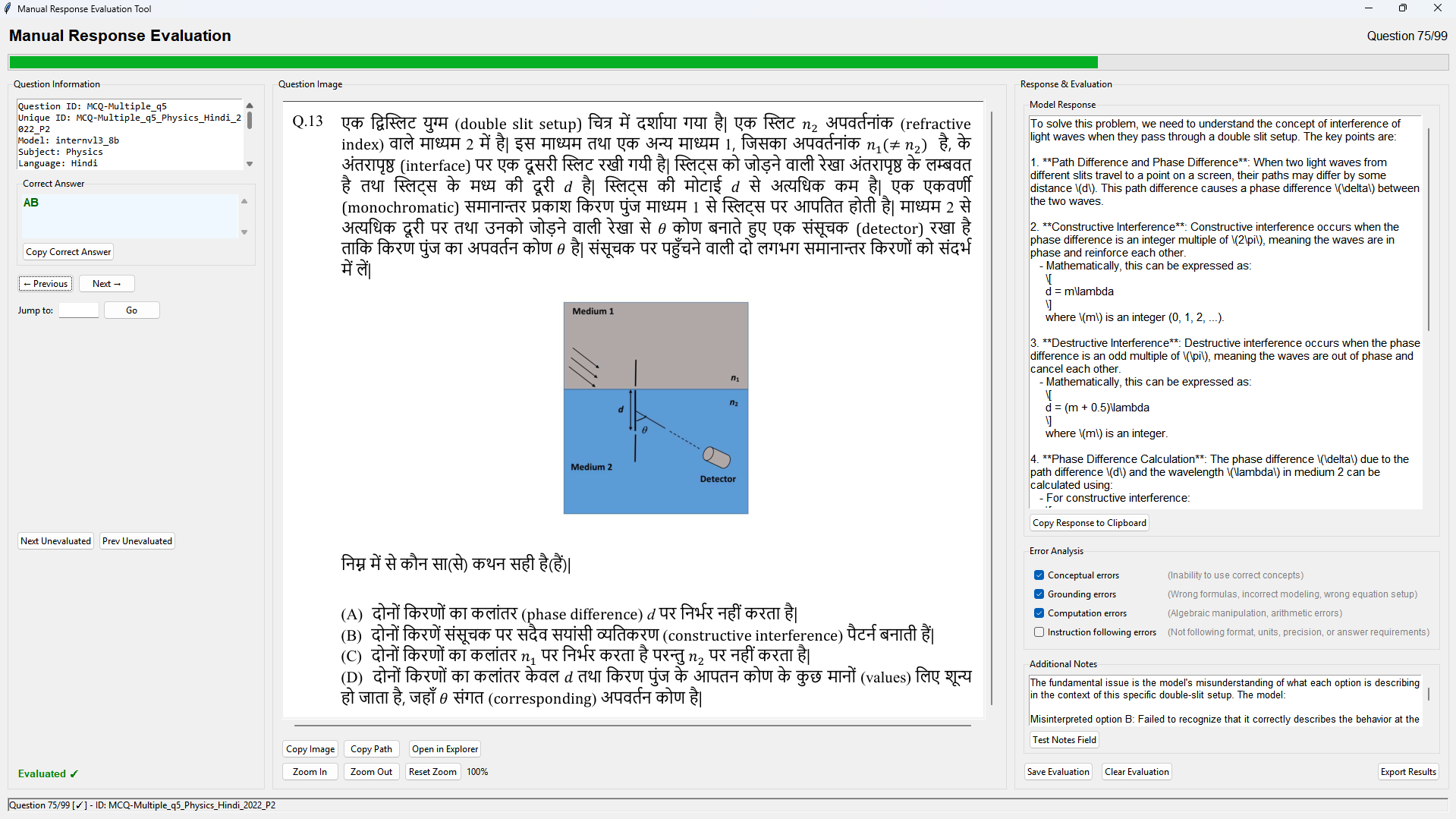}
    \caption{Our manual verification utility loads the question, correct answer, and the model's response. The human annotator evaluates all presented information and votes on the kind(s) of errors found.}
    \label{fig:manual_verification_tool}
\end{figure}

\textbf{Manual Evaluation Interface.} Our response evaluation tool (Figure~\ref{fig:manual_verification_tool}) facilitates systematic manual analysis of model responses across four error dimensions: conceptual errors, grounding errors, computation errors, and instruction-following errors. The interface displays question metadata and correct answers, renders question images with zoom and navigation controls, and presents model responses alongside evaluation forms. The tool supports checkpoint-based evaluation sessions, allowing annotators to pause and resume lengthy evaluation processes. Additional features include clipboard functionality for copying images and responses, navigation shortcuts to unevaluated questions, and export capabilities for generating analysis-ready CSV files. This framework enabled our comprehensive error analysis across 400 manually reviewed questions, providing the granular insights presented in Table~\ref{tab:human_eval_quality}.

\section{Studying Failure Cases}
\label{sec:failure-cases}

Now, we look at a few cherry-picked failure cases to understand where VLMs make mistakes while solving \jee. We present two examples from models with different profiles: the open-source sub-10B InternVL3 10B, and state-of-the-art proprietary Gemini 2.5 Pro. Following is an analysis of where these models make mistakes to better understand their limitations.

\subsection{Example 1: Conceptual and Grounding Errors}

Following is an example of a factual hallucination or mathematical reasoning error. The model incorrectly evaluates the validity of a clearly solvable algebraic equation. Even though the equation is standard and solvable, the model confidently declares it invalid, indicating a lack of basic symbolic reasoning or over-reliance on heuristics rather than step-by-step computation. It reflects failures in math grounding despite surface-level fluency.

\begin{tcolorbox}[
    breakable,
    enhanced,
    colback=white,
    colframe=black,
    title=Numerical\_q2\_Maths\_English\_2022\_P1 + Model Response (InternVL3 8B),
    width=\columnwidth,
    fonttitle=\bfseries,
    coltitle=white,
    before upper={\parindent0em},
]

\textbf{Question:}

\begin{tcolorbox}[colback=lightblue, colframe=black, width=\columnwidth]
Let $\alpha$ be a positive real number. Let $f \colon \mathbb{R} \to \mathbb{R}$ and $g \colon (\alpha, \infty) \to \mathbb{R}$ be the functions defined by
\[
f(x) = \sin\left(\frac{\pi x}{12}\right)
\]
\[
\text{and} \quad g(x) = \frac{2 \log_e \left( \sqrt{x} - \sqrt{\alpha} \right)}{\log_e \left( e^{\sqrt{x}} - e^{\sqrt{\alpha}} \right)}.
\]
Then the value of $\displaystyle \lim_{x \to \alpha^+} f(g(x))$ is \underline{\hspace{2cm}}.
\end{tcolorbox}

\vspace{0.3cm}

\textbf{Model Output (InternVL3 8B):}

\begin{tcolorbox}[colback=lightred, colframe=black, width=\columnwidth, title=Steps 1–2]
To solve for $\lim_{x \to 0^+} g(f(x))$, we need to evaluate the composition of functions as $x$ approaches zero from the positive side.

\textbf{Step 1: Evaluate $f(x)$ as $x \to 0^+$}
\[
f(x) = \sin\left( \frac{\pi x}{12} \right)
\]
\[
\Rightarrow f(0) = \sin(0) = 0
\]
\textbf{Step 2: Evaluate $g(f(x))$}
\[
g(x) = \frac{2 \log_6(\sqrt{x} - \sqrt[4]{x})}{\log_8(e^x - e^{-x})}
\]

\end{tcolorbox}

\begin{tcolorbox}[colback=lightred, colframe=black, width=\columnwidth, title=Steps 2–6]
\[
\Rightarrow \lim_{x \to 0^+} g(f(x)) = g(0)
\]

\textbf{Step 3: Simplify $g(0)$}
\[
g(0) = \frac{2 \log_6(0)}{\log_8(0)} \quad 
\]
\[
\wrongtag{(Wrong formula usage)}
\]

\textbf{Step 4: Use Taylor Series Approximations}
\[
e^x \approx 1 + x, \quad e^{-x} \approx 1 - x 
\]
\[
\Rightarrow e^x - e^{-x} \approx 2x
\]
\[
\log_8(e^x - e^{-x}) \approx \log_8(2x) = \frac{\ln(2x)}{\ln(8)}
\]
\textbf{Step 5: Simplify the numerator}
\[
\sqrt{x} - \sqrt[4]{x} = x^{1/2} - x^{1/4} 
\]
\[
= x^{1/4}(x^{1/4} - 1) \approx -x^{1/4}
\]
\[
\log_6(\sqrt{x} - \sqrt[4]{x}) \approx \log_6(-x^{1/4}) 
\]
\[
\quad \wrongtag{(Wrong assumption)}
\]

\textbf{Step 6: Combine numerator and denominator}
\[
g(0) \approx \frac{\frac{\ln(x)}{2\ln(6)}}{\frac{\ln(x)}{3\ln(2)}} = \frac{3\ln(2)}{2\ln(6)} = \boxed{6}
\]

\vspace{0.2cm}
\wrongtag{Wrong answer: Interpreted limit at 0 instead of $\alpha$ and used incorrect domains.}
\end{tcolorbox}

\vspace{0.3cm}


\begin{tcolorbox}[colback=lightgreen, colframe=black, title=Final Evaluation]
\textbf{Predicted:} \fbox{6} \\
\textbf{Ground Truth:} \fbox{0.5} \\
\textbf{Marking:} \textcolor{red}{Incorrect due to conceptual and grounding errors}
\end{tcolorbox}

\end{tcolorbox}

\subsection{Language Control Failure}

The model unexpectedly responds in German, despite the prompt being in English and the input image in Hindi. This illustrates a language control failure, where the model drifts from the user's intended language — a critical issue in real-world deployments where understanding cultural cues from the supplied prompt is important before adapting model outputs to fit the user.

\begin{tcolorbox}[
    breakable,
    enhanced,
    colback=white,
    colframe=black,
    coltitle=white,
    title=Numerical\_q4\_Physics\_Hindi\_2021\_P2 + Model Response (Gemini 2.5 Pro),
    width=\columnwidth,
    fonttitle=\bfseries,
    before upper={\parindent0em},
]

\textbf{Question Image:}
\begin{center}
    \includegraphics[width=\columnwidth]{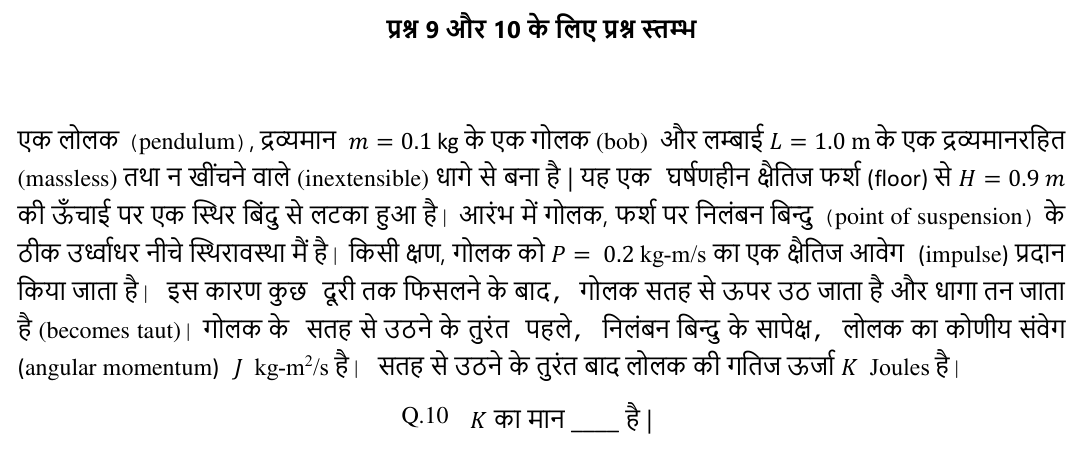}
\end{center}

\vspace{0.5cm}

\textbf{Correct Answer:} \fbox{0.16} \hfill \newline
\textbf{Predicted Answer:} \fbox{0.162} \hfill \newline
\textbf{Is Correct:} \textcolor{red}{\textbf{No}}

\vspace{0.3cm}

\begin{tcolorbox}[colback=lightred, colframe=black, title=Model's Response (Language: German)]
\textbf{Analyse des Problems (Problem Analysis)}

Basierend auf dem Hindi-Text lassen sich die folgenden physikalischen Parameter und Bedingungen extrahieren:

\begin{itemize}
    \item Masse des Pendelkörpers (bob), $m = 0.1$ kg
    \item Länge des Fadens, $L = 1.0$ m (masselos, nicht dehnbar)
    \item Höhe des Aufhängepunktes über dem Boden, $H = 0.9$ m
    \item Boden: Reibungslose horizontale Fläche
    \item Anfangszustand: Der Pendelkörper ruht unter dem Aufhängepunkt auf dem Boden ($L > H$)
    \item Ereignis: Ein Impuls $P = 0.2$ kg·m/s wirkt horizontal auf den Körper
\end{itemize}

\vspace{0.2cm}

\end{tcolorbox}

\begin{tcolorbox}[colback=lightred, colframe=black, width=\columnwidth, title=Steps 1-4]
\begin{itemize}
    \item Gesucht: Kinetische Energie $K$ unmittelbar nachdem der Faden straff wird
\end{itemize}

\vspace{0.2cm}
\textbf{Schritt 1: Anfangsgeschwindigkeit}
\[
v_0 = \frac{P}{m} = \frac{0.2}{0.1} = 2.0 \, \text{m/s}
\]

\vspace{0.2cm}
\textbf{Schritt 2: Geometrie beim Straffen des Fadens}

Ein rechtwinkliges Dreieck mit:
\[
\text{Vertikale: } H = 0.9 \quad
\]
\[
\text{Hypotenuse: } L = 1.0
\]
\[
\Rightarrow \cos(\theta) = \frac{0.9}{1.0} = 0.9
\]
\textbf{Schritt 3: Geschwindigkeit senkrecht zum Faden}

Nur der senkrechte Anteil bleibt:
\[
v_f = v_0 \cdot \cos(\theta) = 2.0 \cdot 0.9 = 1.8 \, \text{m/s}
\]

\vspace{0.2cm}
\textbf{Schritt 4: Kinetische Energie}
\[
K = \frac{1}{2} m v_f^2 = \frac{1}{2} \cdot 0.1 \cdot (1.8)^2 
\]
\[
= 0.05 \cdot 3.24 = 0.162 \, \text{Joule}
\]

\vspace{0.2cm}
\[
\boxed{0.162}
\]

\wrongtag{Note: Although the calculation is correct, it does not match the accepted correct value \textbf{0.16}. The predicted answer is slightly over the acceptable threshold.}
\end{tcolorbox}

\begin{tcolorbox}[colback=lightgreen, colframe=black, title=Final Evaluation]
\textbf{Predicted:} \fbox{0.162} \\
\textbf{Ground Truth:} \fbox{0.16} \\
\textbf{Marking:} \textcolor{red}{Incorrect due to precision threshold}
\end{tcolorbox}

\end{tcolorbox}

\section{JEE Advanced Scoring System}
\label{sec:scoring-system}

\begin{table*}[htbp]
\centering
\begin{tabular}{|l|l|l|}
\hline
\textbf{Question Type} & \textbf{Response Type} & \textbf{Marks} \\
\hline
\multirow{3}{*}{MCQ (Single Correct)} 
    & Correct & +3 \\
    & Wrong   & -1 \\
    & Unanswered & 0 \\
\hline
\multirow{6}{*}{MCQ (Multiple Correct)} 
    & All correct options chosen & +4 \\
    & 3 correct options chosen (of 4) & +3 \\
    & 2 correct options chosen (of 3+) & +2 \\
    & 1 correct option chosen (of 2+) & +1 \\
    & Wrong / Any incorrect option chosen & -2 \\
    & Unanswered & 0 \\
\hline
\multirow{2}{*}{Numerical Answer Type} 
    & Correct & +4 \\
    & Wrong / Unanswered & 0 \\
\hline
\multirow{3}{*}{Matching Type} 
    & Correct & +4 \\
    & Wrong & -1 \\
    & Unanswered & 0 \\
\hline
\end{tabular}
\caption{JEE Advanced Marking Scheme}
\label{tab:marking_scheme}
\end{table*}

Table~\ref{tab:marking_scheme} summarizes the marking scheme for JEE Advanced 2025. The questions involve a mix of positive and negative rewards, which adds to the difficulty of the exam and \jee.

\section{Datasheet}

This datasheet is based on~\citet{gebru2021datasheets}. For the code and LaTeX template, check \href{https://github.com/AudreyBeard/Datasheets-for-Datasets-Template}{their GitHub repo}.

\begin{mdframed}[linecolor=\sectioncolor]
\section*{\textcolor{\sectioncolor}{
    MOTIVATION
}}
\end{mdframed}

    \textcolor{\sectioncolor}{\textbf{
    For what purpose was the dataset created?
    }
    Was there a specific task in mind? Was there
    a specific gap that needed to be filled? Please provide a description.
    } \\
    \jee~is constructed for evaluating state-of-the-art vision-language models (VLMs) on STEM reasoning. he dataset was developed to advance evaluation methodologies for vision-language models in multilingual and multimodal scientific reasoning contexts. \\
    
    \textcolor{\sectioncolor}{\textbf{
    Who created this dataset (e.g., which team, research group) and on behalf
    of which entity (e.g., company, institution, organization)?
    }
    } \\
    The dataset was created by the authors, Arka Mukherjee and Shreya Ghosh, as part of the NLP research group at IIT Bhubaneswar. \\
    
    \textcolor{\sectioncolor}{\textbf{
    What support was needed to make this dataset?
    }
    (e.g.who funded the creation of the dataset? If there is an associated
    grant, provide the name of the grantor and the grant name and number, or if
    it was supported by a company or government agency, give those details.)
    } \\
    Anonymized for peer review \\
    
    \textcolor{\sectioncolor}{\textbf{
    Any other comments?
    }} \\
    N/A \\

\begin{mdframed}[linecolor=\sectioncolor]
\section*{\textcolor{\sectioncolor}{
    COMPOSITION
}}
\end{mdframed}
    \textcolor{\sectioncolor}{\textbf{
    What do the instances that comprise the dataset represent (e.g., documents,
    photos, people, countries)?
    }
    Are there multiple types of instances (e.g., movies, users, and ratings;
    people and interactions between them; nodes and edges)? Please provide a
    description.
    } \\
    Each instance is a JEE Advanced examination question comprising a question image (containing mathematical notation, diagrams, and text in Hindi or English), metadata (year, paper, subject, question type), and ground truth answers verified through multi-source consensus. \\
    
    \textcolor{\sectioncolor}{\textbf{
    How many instances are there in total (of each type, if appropriate)?
    }
    } \\
    The dataset contains 1,460 questions spanning seven years (2019-2025) of JEE Advanced examinations, distributed across Chemistry (492 questions, 33.7\%), Mathematics (492 questions, 33.7\%), and Physics (476 questions, 32.6\%). Question types include Numerical (652 instances, 44.7\%), MCQ-Multiple (412 instances, 28.2\%), MCQ-Single (396 instances, 27.1\%), and Matching (112 instances). \\
    
    \textcolor{\sectioncolor}{\textbf{
    Does the dataset contain all possible instances or is it a sample (not
    necessarily random) of instances from a larger set?
    }
    If the dataset is a sample, then what is the larger set? Is the sample
    representative of the larger set (e.g., geographic coverage)? If so, please
    describe how this representativeness was validated/verified. If it is not
    representative of the larger set, please describe why not (e.g., to cover a
    more diverse range of instances, because instances were withheld or
    unavailable).
    } \\
    This dataset is a subset of all publicly available JEE Advanced questions from 2007-2025. We select editions from 2019-2025 as they specifically incorporates the multilingual aspect (the exam was offered only in English till 2018). \\
    
    \textcolor{\sectioncolor}{\textbf{
    What data does each instance consist of?
    }
    “Raw” data (e.g., unprocessed text or images) or features? In either case,
    please provide a description.
    } \\
    Each instance contains: (1) a high-resolution image of the question extracted from official PDF papers, (2) structured metadata including year, paper number, language, subject, question type, and unique identifier, (3) ground truth answers verified through majority voting across six independent sources, and (4) confidence scores for answer verification. \\
    
    \textcolor{\sectioncolor}{\textbf{
    Is there a label or target associated with each instance?
    }
    If so, please provide a description.
    } \\
    Yes, each instance has an associated ground truth answer obtained through multi-source verification. For MCQ questions, labels are option letters (A, B, C, D) or combinations thereof. For numerical questions, labels are precise numerical values. \\
    
    \textcolor{\sectioncolor}{\textbf{
    Is any information missing from individual instances?
    }
    If so, please provide a description, explaining why this information is
    missing (e.g., because it was unavailable). This does not include
    intentionally removed information, but might include, e.g., redacted text.
    } \\
    No essential information is missing from individual instances. All questions include images, metadata, and verified answers. Some instances may lack official answer keys, but these are compensated through our multi-source verification approach using independent coaching institute solutions. \\
    
    \textcolor{\sectioncolor}{\textbf{
    Are relationships between individual instances made explicit (e.g., users’
    movie ratings, social network links)?
    }
    If so, please describe how these relationships are made explicit.
    } \\
    Instances are grouped by examination metadata (year, paper, subject) and question type. Questions from the same examination paper may share thematic connections, but each instance is designed to be independently evaluable. No explicit dependency relationships exist between questions. \\
    
    \textcolor{\sectioncolor}{\textbf{
    Are there recommended data splits (e.g., training, development/validation,
    testing)?
    }
    If so, please provide a description of these splits, explaining the
    rationale behind them.
    } \\
    No pre-defined training/testing splits are provided, as this is an evaluation-only dataset. Users may create custom splits based on years, subjects, or question types, depending on their evaluation needs. We recommend using temporal splits (e.g., 2019-2023 vs 2024-2025) for contamination-aware evaluation. \\
    
    \textcolor{\sectioncolor}{\textbf{
    Are there any errors, sources of noise, or redundancies in the dataset?
    }
    If so, please provide a description.
    } \\
    Questions with answer disagreement below 60\% consensus were flagged for manual review. Only 30 out of 1,476 initial questions exhibited such disagreements. Such questions were primarily numericals, where official keys accept ranges while coaching institutes provide point values. No significant redundancies exist as each question represents a unique examination item. \\
    
    \textcolor{\sectioncolor}{\textbf{
    Is the dataset self-contained, or does it link to or otherwise rely on
    external resources (e.g., websites, tweets, other datasets)?
    }
    If it links to or relies on external resources, a) are there guarantees
    that they will exist, and remain constant, over time; b) are there official
    archival versions of the complete dataset (i.e., including the external
    resources as they existed at the time the dataset was created); c) are
    there any restrictions (e.g., licenses, fees) associated with any of the
    external resources that might apply to a future user? Please provide
    descriptions of all external resources and any restrictions associated with
    them, as well as links or other access points, as appropriate.
    } \\
    The dataset is self-contained. \\
    
    \textcolor{\sectioncolor}{\textbf{
    Does the dataset contain data that might be considered confidential (e.g.,
    data that is protected by legal privilege or by doctor-patient
    confidentiality, data that includes the content of individuals’ non-public
    communications)?
    }
    If so, please provide a description.
    } \\
    No, the dataset contains only publicly available examination questions from official JEE Advanced papers. \\
    
    \textcolor{\sectioncolor}{\textbf{
    Does the dataset contain data that, if viewed directly, might be offensive,
    insulting, threatening, or might otherwise cause anxiety?
    }
    If so, please describe why.
    } \\
    No \\
    
    \textcolor{\sectioncolor}{\textbf{
    Does the dataset relate to people?
    }
    If not, you may skip the remaining questions in this section.
    } \\
    No \\
    
    \textcolor{\sectioncolor}{\textbf{
    Does the dataset identify any subpopulations (e.g., by age, gender)?
    }
    If so, please describe how these subpopulations are identified and
    provide a description of their respective distributions within the dataset.
    } \\
    No \\
    
    \textcolor{\sectioncolor}{\textbf{
    Is it possible to identify individuals (i.e., one or more natural persons),
    either directly or indirectly (i.e., in combination with other data) from
    the dataset?
    }
    If so, please describe how.
    } \\
    No \\
    
    \textcolor{\sectioncolor}{\textbf{
    Does the dataset contain data that might be considered sensitive in any way
    (e.g., data that reveals racial or ethnic origins, sexual orientations,
    religious beliefs, political opinions or union memberships, or locations;
    financial or health data; biometric or genetic data; forms of government
    identification, such as social security numbers; criminal history)?
    }
    If so, please provide a description.
    } \\
    No \\
    
    \textcolor{\sectioncolor}{\textbf{
    Any other comments?
    }} \\
    N/A
\begin{mdframed}[linecolor=\sectioncolor]
\section*{\textcolor{\sectioncolor}{
    COLLECTION
}}
\end{mdframed}

    \textcolor{\sectioncolor}{\textbf{
    How was the data associated with each instance acquired?
    }
    Was the data directly observable (e.g., raw text, movie ratings),
    reported by subjects (e.g., survey responses), or indirectly
    inferred/derived from other data (e.g., part-of-speech tags, model-based
    guesses for age or language)? If data was reported by subjects or
    indirectly inferred/derived from other data, was the data
    validated/verified? If so, please describe how.
    } \\
    Question images were directly extracted from official JEE Advanced PDF papers using custom annotation tools. Ground truth answers were collected from six independent sources (five coaching institutes plus official keys, detailed in Sec~\ref{sec:data-stat}) and verified through majority voting consensus. All source materials are publicly available through official examination archives. \\
    
    \textcolor{\sectioncolor}{\textbf{
    Over what timeframe was the data collected?
    }
    Does this timeframe match the creation timeframe of the data associated
    with the instances (e.g., recent crawl of old news articles)? If not,
    please describe the timeframe in which the data associated with the
    instances was created. Finally, list when the dataset was first published.
    } \\
    Data collection was conducted between June to July 2025. \\
    
    \textcolor{\sectioncolor}{\textbf{
    What mechanisms or procedures were used to collect the data (e.g., hardware
    apparatus or sensor, manual human curation, software program, software
    API)?
    }
    How were these mechanisms or procedures validated?
    } \\
    We developed custom software tools for PDF processing, question extraction, and answer verification (Appendix~\ref{sec:manual-collection-software}). \\
    
    \textcolor{\sectioncolor}{\textbf{
    What was the resource cost of collecting the data?
    }
    (e.g. what were the required computational resources, and the associated
    financial costs, and energy consumption - estimate the carbon footprint.)
    } \\
    Computational costs were minimal, primarily involving PDF processing and image extraction. \\
    
    \textcolor{\sectioncolor}{\textbf{
    If the dataset is a sample from a larger set, what was the sampling
    strategy (e.g., deterministic, probabilistic with specific sampling
    probabilities)?
    }
    } \\
    We select questions from the past seven editions of JEE Advanced (2019-2025) as the exam has been offered in Hindi since 2019. The sole criterion for this sample was to ensure an equitable distribution of questions in the two languages. \\
    
    \textcolor{\sectioncolor}{\textbf{
    Who was involved in the data collection process (e.g., students,
    crowdworkers, contractors) and how were they compensated (e.g., how much
    were crowdworkers paid)?
    }
    } \\
    Data collection was performed by the authors.  \\
    
    \textcolor{\sectioncolor}{\textbf{
    Were any ethical review processes conducted (e.g., by an institutional
    review board)?
    }
    If so, please provide a description of these review processes, including
    the outcomes, as well as a link or other access point to any supporting
    documentation.
    } \\
    No formal institutional review board process was conducted as the dataset consists entirely of publicly available examination materials with no human subjects involved. \\
    
    \textcolor{\sectioncolor}{\textbf{
    Does the dataset relate to people?
    }
    If not, you may skip the remainder of the questions in this section.
    } \\
    N/A \\
    
    \textcolor{\sectioncolor}{\textbf{
    Did you collect the data from the individuals in question directly, or
    obtain it via third parties or other sources (e.g., websites)?
    }
    } \\
    All questions were collected from the public JEE Advanced archive. Answers have been sourced from the official archives of the respective coaching institutes. \\
    
    \textcolor{\sectioncolor}{\textbf{
    Were the individuals in question notified about the data collection?
    }
    If so, please describe (or show with screenshots or other information) how
    notice was provided, and provide a link or other access point to, or
    otherwise reproduce, the exact language of the notification itself.
    } \\
    No, since all data has been publicly released. \\
    
    \textcolor{\sectioncolor}{\textbf{
    Did the individuals in question consent to the collection and use of their
    data?
    }
    If so, please describe (or show with screenshots or other information) how
    consent was requested and provided, and provide a link or other access
    point to, or otherwise reproduce, the exact language to which the
    individuals consented.
    } \\
    N/A \\
    
    \textcolor{\sectioncolor}{\textbf{
    If consent was obtained, were the consenting individuals provided with a
    mechanism to revoke their consent in the future or for certain uses?
    }
     If so, please provide a description, as well as a link or other access
     point to the mechanism (if appropriate)
    } \\
    N/A \\
    
    \textcolor{\sectioncolor}{\textbf{
    Has an analysis of the potential impact of the dataset and its use on data
    subjects (e.g., a data protection impact analysis)been conducted?
    }
    If so, please provide a description of this analysis, including the
    outcomes, as well as a link or other access point to any supporting
    documentation.
    } \\
    N/A \\
    
    \textcolor{\sectioncolor}{\textbf{
    Any other comments?
    }} \\
    N/A

\begin{mdframed}[linecolor=\sectioncolor]
\section*{\textcolor{\sectioncolor}{
    PREPROCESSING / CLEANING / LABELING
}}
\end{mdframed}

    \textcolor{\sectioncolor}{\textbf{
    Was any preprocessing/cleaning/labeling of the data
    done(e.g.,discretization or bucketing, tokenization, part-of-speech
    tagging, SIFT feature extraction, removal of instances, processing of
    missing values)?
    }
    If so, please provide a description. If not, you may skip the remainder of
    the questions in this section.
    } \\
    Questions were extracted from PDFs using custom annotation tools with manual verification for accuracy. Images were processed to ensure high resolution and proper cropping. Answer verification involved majority voting across multiple sources with confidence scoring. Questions failing to achieve >60\% consensus were manually reviewed. \\

    \textcolor{\sectioncolor}{\textbf{
    Was the “raw” data saved in addition to the preprocessed/cleaned/labeled
    data (e.g., to support unanticipated future uses)?
    }
    If so, please provide a link or other access point to the “raw” data.
    } \\
    Yes, original PDF files are preserved and referenced. Our annotation tools maintain links to source materials, enabling reproducibility and verification of extraction accuracy. Moreover, all raw PDF files can be accessed through the official JEE Advanced archive. \\

    \textcolor{\sectioncolor}{\textbf{
    Is the software used to preprocess/clean/label the instances available?
    }
    If so, please provide a link or other access point.
    } \\
    We make the \href{https://github.com/ArkaMukherjee0/mmJEE-Eval}{code and dataset} publicly available. \\

    \textcolor{\sectioncolor}{\textbf{
    Any other comments?
    }} \\
    N/A

\begin{mdframed}[linecolor=\sectioncolor]
\section*{\textcolor{\sectioncolor}{
    USES
}}
\end{mdframed}

    \textcolor{\sectioncolor}{\textbf{
    Has the dataset been used for any tasks already?
    }
    If so, please provide a description.
    } \\
    This dataset was created for evaluating vision-language models on authentic multimodal scientific reasoning tasks. It has been used to evaluate five state-of-the-art VLMs, including Gemini 2.5 Pro, OpenAI o3, and smaller open-source models. \\

    \textcolor{\sectioncolor}{\textbf{
    Is there a repository that links to any or all papers or systems that use the dataset?
    }
    If so, please provide a link or other access point.
    } \\
    A general list of the leaderboard and latest news can be viewed at this \href{https://mmjee-eval.github.io}{HTTPS URL}. \\

    \textcolor{\sectioncolor}{\textbf{
    What (other) tasks could the dataset be used for?
    }
    } \\
    The dataset could be used for multilingual mathematical reasoning research, educational AI development, cross-cultural evaluation of AI systems, multimodal representation learning, and development of specialized educational assessment tools for STEM subjects. \\

    \textcolor{\sectioncolor}{\textbf{
    Is there anything about the composition of the dataset or the way it was
    collected and preprocessed/cleaned/labeled that might impact future uses?
    }
    For example, is there anything that a future user might need to know to
    avoid uses that could result in unfair treatment of individuals or groups
    (e.g., stereotyping, quality of service issues) or other undesirable harms
    (e.g., financial harms, legal risks) If so, please provide a description.
    Is there anything a future user could do to mitigate these undesirable
    harms?
    } \\
    Users should be aware that this dataset reflects the specific cultural and educational context of Indian competitive examinations. Performance gaps between languages or subjects might reflect training data biases rather than inherent model capabilities. Care should be taken when generalizing results to other educational contexts or populations. \\

    \textcolor{\sectioncolor}{\textbf{
    Are there tasks for which the dataset should not be used?
    }
    If so, please provide a description.
    } \\
    We advise against including our dataset in the pre-training of VLMs to ensure fair evaluation on future models. This dataset should not be used for commercial assessment tools without proper licensing, for creating competing examination systems that might disadvantage students, or for any purpose that could harm the integrity of the JEE Advanced examination process. \\

    \textcolor{\sectioncolor}{\textbf{
    Any other comments?
    }} \\
    N/A

\begin{mdframed}[linecolor=\sectioncolor]
\section*{\textcolor{\sectioncolor}{
    DISTRIBUTION
}}
\end{mdframed}

    \textcolor{\sectioncolor}{\textbf{
    Will the dataset be distributed to third parties outside of the entity
    (e.g., company, institution, organization) on behalf of which the dataset
    was created?
    }
    If so, please provide a description.
    } \\
    We make the \href{https://github.com/ArkaMukherjee0/mmJEE-Eval}{code} publicly available. The dataset has been publicly released on \href{https://www.huggingface.co/datasets/ArkaMukherjee/mmJEE-Eval}{Hugging Face}. \\

    \textcolor{\sectioncolor}{\textbf{
    How will the dataset be distributed (e.g., tarball on website, API,
    GitHub)?
    }
    Does the dataset have a digital object identifier (DOI)?
    } \\
    Anonymized for peer review \\

    \textcolor{\sectioncolor}{\textbf{
    When will the dataset be distributed?
    }
    } \\
    The dataset is publicly available from Hugging Face from the date of publication. \\

    \textcolor{\sectioncolor}{\textbf{
    Will the dataset be distributed under a copyright or other intellectual
    property (IP) license, and/or under applicable terms of use (ToU)?
    }
    If so, please describe this license and/or ToU, and provide a link or other
    access point to, or otherwise reproduce, any relevant licensing terms or
    ToU, as well as any fees associated with these restrictions.
    } \\
    The dataset is being released under the MIT license permitting academic and research use. Since source materials are public examination papers, no additional licensing restrictions apply beyond standard academic attribution requirements. \\

    \textcolor{\sectioncolor}{\textbf{
    Have any third parties imposed IP-based or other restrictions on the data
    associated with the instances?
    }
    If so, please describe these restrictions, and provide a link or other
    access point to, or otherwise reproduce, any relevant licensing terms, as
    well as any fees associated with these restrictions.
    } \\
    No third-party IP restrictions apply, as all source materials are publicly available examination papers. \\

    \textcolor{\sectioncolor}{\textbf{
    Do any export controls or other regulatory restrictions apply to the
    dataset or to individual instances?
    }
    If so, please describe these restrictions, and provide a link or other
    access point to, or otherwise reproduce, any supporting documentation.
    } \\
    None \\

    \textcolor{\sectioncolor}{\textbf{
    Any other comments?
    }} \\
    N/A \\

\begin{mdframed}[linecolor=\sectioncolor]
\section*{\textcolor{\sectioncolor}{
    MAINTENANCE
}}
\end{mdframed}

    \textcolor{\sectioncolor}{\textbf{
    Who is supporting/hosting/maintaining the dataset?
    }
    } \\
    The dataset is supported by the SPARC  \\

    \textcolor{\sectioncolor}{\textbf{
    How can the owner/curator/manager of the dataset be contacted (e.g., email
    address)?
    }
    } \\
    The authors can be reached at the following email addresses:
    \begin{itemize}
        \item Arka Mukherjee: \href{mailto:arka.mukherjee078@gmail.com}{arka.mukherjee078@gmail\\.com}
        \item Shreya Ghosh: \href{mailto:shreya@iitbbs.ac.in}{shreya@iitbbs.ac.in}
    \end{itemize} 

    \textcolor{\sectioncolor}{\textbf{
    Is there an erratum?
    }
    If so, please provide a link or other access point.
    } \\
    N/A \\

    \textcolor{\sectioncolor}{\textbf{
    Will the dataset be updated (e.g., to correct labeling errors, add new
    instances, delete instances)?
    }
    If so, please describe how often, by whom, and how updates will be
    communicated to users (e.g., mailing list, GitHub)?
    } \\
    Yes, in the June of every year, approximately 190 JEE Advanced questions will be added to the dataset, with leaderboard results publicized on \href{https://mmjee-eval.github.io}{our website}. \\

    \textcolor{\sectioncolor}{\textbf{
    If the dataset relates to people, are there applicable limits on the
    retention of the data associated with the instances (e.g., were individuals
    in question told that their data would be retained for a fixed period of
    time and then deleted)?
    }
    If so, please describe these limits and explain how they will be enforced.
    } \\
    N/A \\

    \textcolor{\sectioncolor}{\textbf{
    Will older versions of the dataset continue to be
    supported/hosted/maintained?
    }
    If so, please describe how. If not, please describe how its obsolescence
    will be communicated to users.
    } \\
    Yes \\

    \textcolor{\sectioncolor}{\textbf{
    If others want to extend/augment/build on/contribute to the dataset, is
    there a mechanism for them to do so?
    }
    If so, please provide a description. Will these contributions be
    validated/verified? If so, please describe how. If not, why not? Is there a
    process for communicating/distributing these contributions to other users?
    If so, please provide a description.
    } \\
    TBD \\

    \textcolor{\sectioncolor}{\textbf{
    Any other comments?
    }} \\
    N/A

\end{document}